\documentclass[a4paper,fleqn]{cas-dc}

\usepackage{multirow}
\usepackage{booktabs}
\usepackage[authoryear]{natbib}
\usepackage{placeins}
\usepackage{cuted}
\usepackage{caption}

\def\tsc#1{\csdef{#1}{\textsc{\lowercase{#1}}\xspace}}
\tsc{WGM}
\tsc{QE}
\usepackage{algorithm}
\usepackage{algpseudocode}

\begin{document}
	\let\WriteBookmarks\relax
	\def\floatpagepagefraction{1}
	\def\textpagefraction{.001}
	
	\shorttitle{Dual-Branch Underwater Image Enhancement and Object Detection}    
	
	\shortauthors{L. Cao et al.}  
	
	\title [mode = title]{A Dual-Branch Collaborative Framework for Joint Optimization of Underwater Image Enhancement and Object Detection}  

\author[1]{Liyuan Cao}
\ead{2451452455@qq.com}
\credit{Investigation, Writing -- original draft}

\author[1]{Zheng Liu}[orcid=0000-0002-9209-6622]
\cormark[1]
\ead{lz6949505@163.com}
\credit{Supervision, Writing -- review}

\author[1]{Guanghao Liao}[orcid=0009-0005-0966-5839]
\ead{liaoguanghao0826@163.com}
\credit{Methodology, Writing -- editing}

\author[1]{Yonghui Yang}
\ead{Yangyh2636688@163.com}
\credit{Resources}

\author[1]{Qi Li}
\ead{liqi_74@163.com}
\credit{Data curation}

\affiliation[1]{
	organization={School of Electronic and Information Engineering, University of Science and Technology Liaoning},
    addressline={No.189 Qianshan Middle Road, Lishan District},
	city={Anshan},
	postcode={114051},
	state={Liaoning},
	country={China}
}

\cortext[1]{Corresponding author}
	
	
	\begin{abstract}
		Due to wavelength dependent light absorption and scattering, underwater images usually suffer from color distortion and blurred details, which limits underwater object detection performance. Existing underwater image enhancement methods mainly focus on visual quality improvement, while it is still difficult to balance enhancement quality, processing efficiency, and downstream detection performance. Therefore, this paper proposes an efficient dual-branch underwater image enhancement framework for object detection. The detail enhancement branch improves brightness and local contrast to recover texture details in dark regions. The color restoration branch uses adaptive compensation to reduce color distortion and improve color gradation. By combining the complementary outputs of the two branches, the proposed framework provides clearer and more informative images for object detection. On the UIEB and EUVP datasets, the proposed method achieves UIQM scores of 2.249 and 2.576. When applied to the YOLOv8 detection task on the URPC dataset, the proposed method improves mAP50 by 2.1\% compared with the baseline. Extensive experiments show that our method improves object detection in complex underwater scenes, while balancing enhancement quality and processing efficiency.
	\end{abstract}
	
	
\begin{highlights}
    \item A dual-branch framework jointly improves underwater details and color fidelity.
    \item The lightweight method avoids complex network training and maintains high efficiency.
    \item The method balances image quality, processing speed, and detection performance.
\end{highlights}

	
	
	\maketitle
	
	\section{Introduction}\label{Introduction}

\begin{figure*}
		\centering
		\includegraphics[width=.9\textwidth]{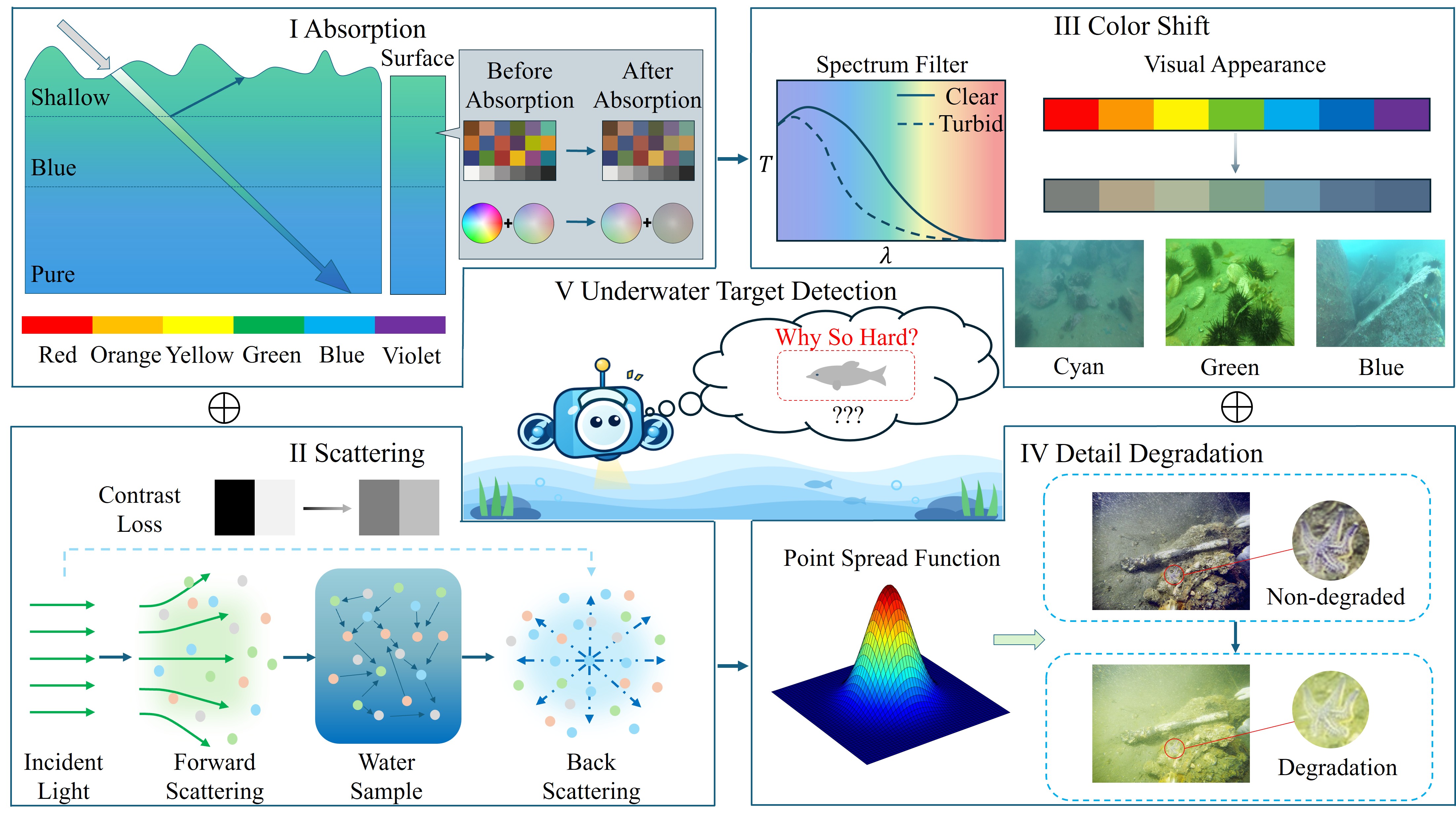}
		\caption{Diagram of a complex underwater environment.}\label{fig1}
	\end{figure*}

	With the growing demand for marine resource development, ecological and environmental monitoring, and intelligent navigation, underwater target detection is playing an increasingly important role in underwater robotic operations, marine observation, and engineering applications. Because light propagation in water involves absorption and scattering, the attenuation rates of light at different wavelengths vary significantly, which can easily cause color shifts in underwater images \citep{Chang2025Mamba}. At the same time, the scattering effect caused by suspended particles in water further reduces image contrast, blurs edge structures, and results in a loss of texture information, as shown in Fig.~\ref{fig1}. These degradation phenomena not only reduce the observability of underwater scenes but also weaken the effective representation of features in the target area, making feature extraction and recognition more difficult, and thereby limiting the accuracy and stability of detection models in complex environments.
	
	Existing methods for underwater image enhancement can be mainly divided into physical model-based methods, model-free methods, and deep learning-based methods. Physical model-based methods restore underwater images by estimating background light and transmission maps, which provides strong physical interpretability. However, their performance highly depends on model assumptions and parameter estimation accuracy, leading to unstable restoration results in complex underwater environments \citep{Song2020Enhancement}. Model-free methods mainly improve the visual quality of underwater images through histogram algorithms \citep{Peng2022HEApproximation} and Retinex algorithms \citep{Zhang2017Retinex}. Although these methods are simple to implement and have low computational cost, they may cause detail distortion and unnatural colors. Deep learning-based methods have achieved significant progress in underwater image enhancement \citep{Sun2024GhostUNet}. However, these methods usually rely on large amounts of high quality training data and high computational resources. Meanwhile, in complex real underwater scenes without reference images, the cross scene generalization ability and enhancement stability of these methods still need to be improved. Some methods may also produce color over correction and texture artifacts, which limits their application in real time deployment and lightweight underwater vision systems.
	
	Most existing research on underwater image enhancement focuses primarily on improving visual quality, and evaluations typically emphasize image quality metrics such as SSIM \citep{Wang2004SSIM}, UIQM \citep{Panetta2016UIQM}, and UCIQE \citep{Yang2015UCIQE}. Previous studies have shown that improved visual quality does not always lead to better downstream object detection performance \citep{Liu2020RealWorld}. Even if the enhanced images show improvements in terms of subjective visual quality or objective evaluation metrics, they may not necessarily provide more effective discriminative features for object detection. Therefore, the design and evaluation of underwater image enhancement methods should not only focus on visual quality, but also consider processing efficiency and their influence on downstream tasks.
    
	To address these problems, we divide underwater image degradation into two complementary aspects: detail degradation and color distortion. Based on the needs of downstream object detection, we propose a dual-branch underwater image enhancement framework. The detail enhancement branch reduces uneven lighting through illumination correction, and improves image layering and texture clarity through brightness adjustment and local detail enhancement. The color restoration branch uses brightness guided color compensation and local contrast optimization to improve color naturalness and overall visual consistency. Through the collaborative dual-branch design, the proposed method improves underwater image quality without complex network training while maintaining high processing efficiency. We apply the enhanced images to YOLOv8 and evaluate the proposed method in terms of image quality, processing efficiency, and detection performance. The results demonstrate its suitability for complex underwater scenes and its potential for practical applications.
	
	The main contributions of this paper are as follows:
	\begin{itemize}
		\item To address detail degradation and color distortion in underwater images, we propose a dual-branch enhancement framework for object detection. The two branches work together to improve image details and restore colors.
		\item We design a lightweight enhancement strategy that requires no complex network training. It improves the visual quality of underwater images while maintaining high processing efficiency, showing strong potential for engineering applications.
		\item We evaluate the proposed method from three aspects: image quality, processing speed, and downstream object detection performance. Experimental results on multiple underwater image datasets show that the proposed method achieves a good balance among these three aspects.
	\end{itemize}
	
	This paper is organized as follows. Section~\ref{Related Work} reviews related studies. Section~\ref{Methods} introduces the proposed method. Sections~\ref{Experimental Results and Analysis} presents the experimental results and analysis. Section~\ref{Conclusion} concludes the paper.
	
	\section{Related Work}\label{Related Work}
	
	\subsection{Image Processing Methods}\label{Image Processing Methods}
	
	In complex imaging environments, images are often affected by insufficient illumination, low contrast, and noise interference, making it difficult to accurately represent target edges, texture structures, and local region information. Improving image visibility, enhancing structural information, and suppressing degrading noise are key research topics in the field of image processing. \citet{He2013GuidedFilter} proposed a directional filtering method that improves filtering computational efficiency while preserving edge structures. Zero-DCE \citep{Guo2020ZeroDCE} improves the brightness of low-light images through curve estimation learning. URetinex-Net \citep{Wu2022URetinexNet} combines Retinex theory to improve image structures and detail representation from the perspectives of illumination and reflection decomposition. The SNR-Aware method \citep{Xu2022SNRAware} uses signal-to-noise ratio distribution to estimate the degradation level of different regions and improve restoration performance in low-light areas. These methods have achieved good performance in brightness restoration, structure preservation, and detail enhancement. However, model driven methods are easily limited by fixed rules, while learning based methods rely heavily on training data and computational resources, and may suffer from insufficient generalization or unstable enhancement performance in cross scene applications.
	
	\subsection{Underwater Image Processing Methods}\label{Underwater Image Processing Methods}
	
	Compared with common imaging scenes, underwater image degradation shows more distinct characteristics. Due to the different absorption and scattering effects of water on different wavelengths of light, underwater images usually suffer from color distortion, brightness attenuation, low contrast, and blurred details. These degradations not only reduce the subjective visual quality of images, but also weaken the ability of subsequent detection algorithms to extract target edges, texture information, and category features. Existing studies mainly focus on three directions: physical model-based methods, model-free methods, and deep learning-based methods. The classical Jaffe-McGlamery simplified underwater imaging model \citep{Jaffe1990UnderwaterModel} provides the theoretical foundation for physical model-based methods. \citet{Li2024LightAttenuation}, \citet{Xiang2023AQSCHE}, and \citet{Zhou2023CrossView} improved underwater image quality from the perspectives of light attenuation modeling, histogram equalization, and deep feature learning, respectively. These methods have achieved certain improvements in color correction, contrast enhancement, and detail restoration. However, physical model-based methods usually rely on imaging assumptions and parameter estimation. Single traditional enhancement methods are difficult to handle multiple degradation factors at the same time, while deep learning-based methods often require large amounts of training data and high computational resources. Therefore, this paper adopts a non-learning collaborative enhancement strategy. It improves degraded underwater images by addressing detail loss and color distortion.
	
	\subsection{Object Detection Methods}\label{Object Detection Methods}
	
	Object detection is a key task in underwater scene perception and target recognition. It is widely applied in underwater robot perception, marine organism monitoring, and underwater target recognition. Object detection methods mainly include two stage detection algorithms and one stage detection algorithms. Two stage methods such as Fast R-CNN \citep{Girshick2015FastRCNN} and Faster R-CNN \citep{Ren2017FasterRCNN} improve detection accuracy and localization stability through region proposal generation and refined classification regression. One-stage methods such as YOLO \citep{Redmon2016YOLO} and SSD \citep{Liu2016SSD} remove the region proposal generation process and directly perform object classification and location regression, achieving faster inference speed and better real-time application potential. YOLOv8 further optimizes feature extraction, feature fusion, and detection head design, achieving a good balance between detection speed and detection accuracy, which makes it suitable as a basic framework for real time underwater object detection tasks. The commonly used CIoU loss in YOLOv8 comprehensively considers overlap area, center point distance, and aspect ratio differences, which can improve bounding box regression accuracy and convergence speed to a certain extent. However, in complex underwater scenes, target scale variation, blurred boundaries, and differences among samples of different quality may still affect localization performance. In this paper, we feed the enhanced underwater images into the detection network. We also optimize the bounding box regression loss to improve target localization and overall detection performance.
	
	\section{Methods}\label{Methods}
	
\begin{figure}
		\centering
		\includegraphics[width=.9\columnwidth]{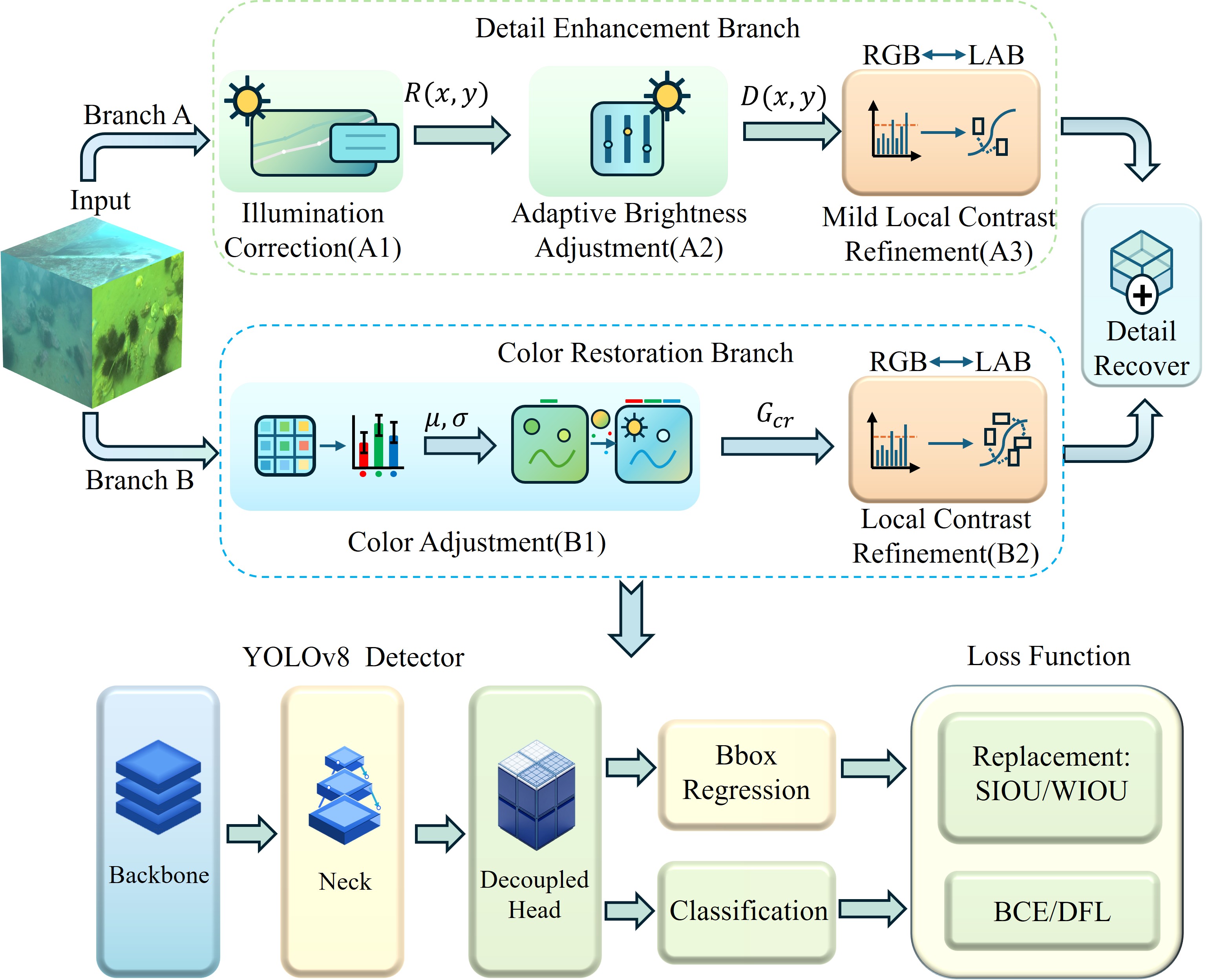}
		\caption{Overall Processing Framework Diagram.}\label{fig2}
	\end{figure}

	Underwater image degradation mainly appears in two forms: detail loss and color distortion. Detail loss blurs object edges and weakens texture information, which makes feature extraction more difficult. Color distortion reduces the contrast between objects and the background, which makes object recognition more difficult. Therefore, this paper builds a detail enhancement branch and a color restoration branch. A weighted fusion strategy combines their complementary information and improves the overall quality of underwater images in a simple and efficient way. The fused images are fed into YOLOv8 for object detection. We also optimize the bounding box regression loss to improve the match between predicted boxes and ground truth boxes. This further improves localization accuracy and overall detection performance. The overall framework of the proposed method is shown in Fig.~\ref{fig2}.
	
	\subsection{Detail Enhancement Branch}
	
\begin{figure}
		\centering
		\includegraphics[width=.9\columnwidth]{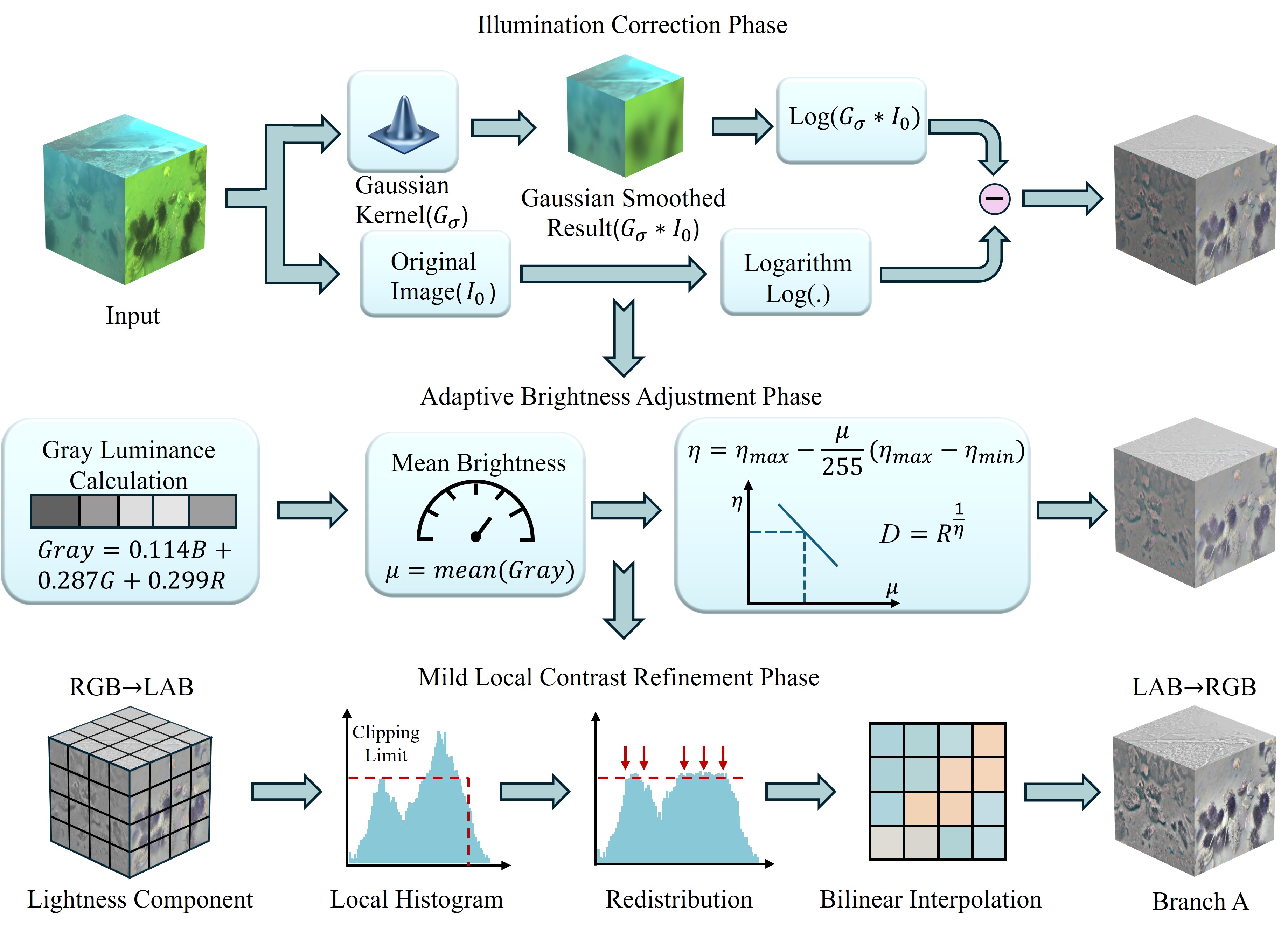}
		\caption{Diagram of the detail enhancement branch.}\label{fig3}
	\end{figure}

	The Detail Enhancement branch comprises three components, including lighting correction, adaptive brightness adjustment, and local contrast enhancement, which optimize the input image in stages to improve its overall visibility and detail rendering. The overall flowchart is shown in Fig.~\ref{fig3}.
	
	To reduce the influence of uneven underwater illumination on image quality, this paper adopts the Single Scale Retinex method for illumination correction of the input image. According to the Retinex theory, an image can be decomposed into illumination and reflection components, where the illumination component mainly contains low frequency information, while the reflection component mainly reflects texture and detail features of the scene and is more related to high frequency information. Based on this idea, the input image is compared with its Gaussian smoothed result in the logarithmic domain. This operation suppresses illumination influence, enhances reflection information, and improves the visibility of dark regions. The Retinex output is formulated as Eq.~\ref{eq1}
	\begin{equation}
		R(x,y)=\log I_0(x,y)-\log\left(G_{\sigma} * I_0\right)(x,y)
		\label{eq1}
	\end{equation}
	Where $G_{\sigma}$ represents a Gaussian function with a standard deviation of $\sigma$.

	After illumination correction, we construct an adaptive adjustment factor based on the overall image brightness. We apply a nonlinear mapping to enhance detail responses, allowing the enhancement strength to adapt to different brightness levels. The mapping process is formulated as Eq.~\ref{eq2}
	\begin{equation}
		D(x,y)=R_n(x,y)^{\frac{1}{\eta}}
		\label{eq2}
	\end{equation}
	Where $\eta$ is an adaptive parameter, which is calculated using Eq.~\ref{eq3}.
	\begin{equation}
		\eta=\eta_{\max}-\frac{\mu}{255}\left(\eta_{\max}-\eta_{\min}\right)
		\label{eq3}
	\end{equation}
    where $\mu$ denotes the average image brightness. The parameters $\eta_{\min}$ and $\eta_{\max}$  only limit the adaptive adjustment range and prevent large fluctuations in mapping strength. They do not serve as the main adjustment parameters.
	
	When the image brightness is low, the mapping process increases the response intensity of dark regions to enhance weak textures and edge structures. When the image brightness is high, the enhancement strength is automatically reduced to avoid local over enhancement and noise amplification, thereby keeping the enhancement process stable and natural.
	
	This paper further introduces a mild local contrast enhancement operation after global brightness adjustment. The mapped result is converted into the LAB color space, and mild local contrast enhancement is applied to the L channel to further improve local brightness differences and texture layering information. As a result, the enhanced image achieves improved overall brightness together with clearer detail representation and richer visual layering.
	
	\subsection{Color Restoration Branch}
	
\begin{figure}
		\centering
		\includegraphics[width=.9\columnwidth]{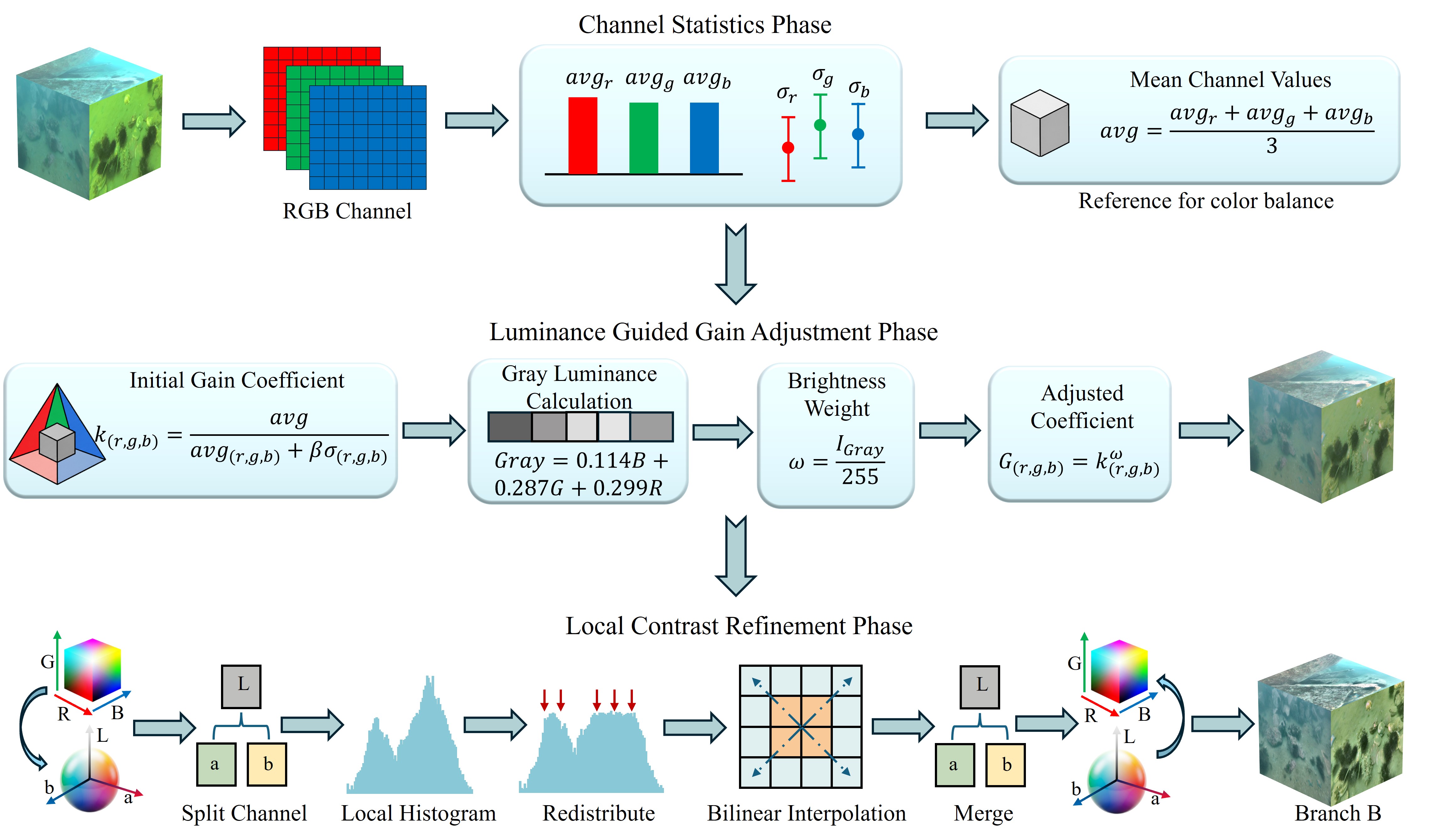}
		\caption{Color Restoration Branch Diagram.}\label{fig4}
	\end{figure}

	The color restoration branch includes channel statistical analysis, brightness guided gain modulation, and local contrast optimization. These components progressively correct the input image to improve color naturalness and visual layering. The overall flow of the branch is shown in Fig.~\ref{fig4}.
	
	The mean value of each color channel is calculated according to Eq.~\ref{eq4}.
	\begin{equation}
		avg_c=\frac{\sum_{x,y} I_{0,c}(x,y)}{N}, \quad c \in \{r,g,b\}
		\label{eq4}
	\end{equation}
	Where $N$ represents the total number of pixels in the image.
	
	Calculate the average of the overall grayscale values using Eq.~\ref{eq5}.
	
	\begin{equation}
		avg = \frac{avg_r + avg_g + avg_b}{3}
		\label{eq5}
	\end{equation}
	
	After obtaining the global statistics, we introduce a small standard deviation correction term into the channel mean to improve the stability of white balance adjustment. We use it to calculate the gain coefficient for each channel. The gain coefficients of different channels are formulated as Eq.~\ref{eq6}
	\begin{equation}
		k_c = \frac{\mathrm{avg}}{\mathrm{avg}_c + \beta\sigma_c + \varepsilon},
		\quad c \in \{r,g,b\}
		\label{eq6}
	\end{equation}
	where $\sigma$ denotes the standard deviation of each channel, and $\varepsilon$ is a small constant that prevents division by zero. The parameter $\beta$ serves as a correction coefficient to limit large fluctuations in the gain of individual channels. We set $\beta$ to 0.1 and keep it unchanged in all experiments. 
    
    Since dark regions in underwater images usually contain low brightness and low confidence color information, strong color compensation may easily cause noise enhancement and false colors.Therefore, we introduce a brightness guided weight to perform spatial adaptive modulation on the color gain. The brightness guided weight is formulated as Eq.~\ref{eq7}.
	\begin{equation}
		\omega(x,y) = \frac{I_{gray}(x,y)}{255}
		\label{eq7}
	\end{equation}
	
	The adaptive variation of channel gain with image brightness is formulated as Eq.~\ref{eq8}.
	\begin{equation}
		G_c(x,y)=k_c^{\,\omega(x,y)},
		\quad c \in \{r,g,b\}
		\label{eq8}
	\end{equation}
	
	The final correction result is formulated as Eq.~\ref{eq9}.
	\begin{equation}
		G_{cr,c}(x,y)=I_{0,c}(x,y)\cdot G_c(x,y),
		\quad c \in \{r,g,b\}
		\label{eq9}
	\end{equation}
	
	When the region brightness is low, the color compensation strength is automatically reduced to suppress dark region noise. In bright regions, the color information is more stable, so the color compensation strength is properly increased to restore attenuated red information and reduce strong blue green color cast. This makes the color distribution more natural and balanced.
	After the above color correction process, image $G_{cr,c}(x,y)$ is converted into the LAB color space, and the L channel is extracted for local layering adjustment. This process redistributes the L channel distribution by limiting the local histogram enhancement strength. The enhanced L channel is then merged with the original a and b channels and converted back to the RGB color space to obtain color restoration result $G_c(x,y)$.
	
	\subsection{Linear weighted fusion}
	
	The processed images $D(x,y)$ and $G_{e}(x,y)$ are linearly weighted and fused according to Eq.~\ref{eq10} to obtain the final image $I_r(x,y)$.
	\begin{equation}
		I_r(x,y) = \partial D(x,y) + (1-\partial)G_e(x,y)
		\label{eq10}
	\end{equation}
	Where $\partial$ represents the weight.
    
	\begin{algorithm}[htbp]
		\caption{Underwater Image Enhancement Algorithm}
		\label{alg1}
		\begin{algorithmic}[1]
			\State \textbf{Input:} Raw underwater image $I_0(x,y)$
			\State \textbf{Output:} Enhanced image $I_r(x,y)$
			\State Estimate illumination component $L(x,y)$ using Gaussian smoothing
			\State Obtain reflectance response $R(x,y)$ by logarithmic difference
			\State Normalize reflectance response $R_n(x,y)$
			\State Calculate luminance mean $\mu$ from $R_n(x,y)$
			\State Determine adaptive adjustment factor $\eta$ based on $\mu$
			\State Perform adaptive nonlinear mapping to obtain $D_1(x,y)$
			\State Apply mild local contrast enhancement to $D_1(x,y)$ in LAB space
			\State Obtain detail-enhanced image $D(x,y)$
			\State Calculate channel means $avg_c$ and standard deviations $\sigma_c$ of $I_0(x,y)$
			\State Calculate global channel average $avg$
			\State Determine statistically constrained channel gains $k_c$
			\State Calculate luminance-guided weight map $\omega(x,y)$ from $I_0(x,y)$
			\State Modulate $k_c$ with $\omega(x,y)$ to obtain adaptive gain map $G_c(x,y)$
			\State Apply adaptive color restoration to obtain $G_{cr,c}(x,y)$
			\State Apply local contrast enhancement to $G_{cr,c}(x,y)$ in LAB space
			\State Obtain color-restored image $G_e(x,y)$
			\State Fuse $D(x,y)$ and $G_e(x,y)$ with weight $\partial$ to obtain final enhanced image $I_r(x,y)$
			\State \textbf{Return} $I_r(x,y)$
		\end{algorithmic}
	\end{algorithm}
	
	\section{Experimental Results and Analysis}\label{Experimental Results and Analysis}
	
	\subsection{Experimental Environment}
	
	The experimental setup details are shown in Table~\ref{tab1}. The experiment was programmed in Python and ran on Windows 11. The model used for the object detection phase was YOLOv8n, with 100 epochs, a batch size of 4, and input image dimensions of $640 \times 640$.
	
	\begin{table}[width=.9\linewidth,cols=2,pos=h]
		\caption{Experimental configuration.}
		\label{tab1}
		\begin{tabular*}{\tblwidth}{@{} CC @{}}
			\toprule
			Experimental Environment & Specification \\
			\midrule
			GPU & NVIDIA GeForce RTX 4070 \\
			CPU & Intel Core i7-10875H \\
			CUDA & 11.1 \\
			PyTorch & 1.8.2 \\
			\bottomrule
		\end{tabular*}
	\end{table}
	
	\subsection{Experimental Dataset}
	
	This paper uses four underwater image datasets-URPC2022, UIEB, EUVP, and DUO-for experimentation.
	
	The UIEB dataset is a commonly used benchmark dataset in underwater image enhancement research. It contains 950 real underwater images, including 890 images with reference images and 60 challenging images. The EUVP dataset contains various complex degradation scenes and can be used to evaluate the enhancement performance of models under different conditions. These two datasets are mainly used for the quantitative evaluation of underwater image enhancement methods to verify enhancement performance under different scenes. We calculate the evaluation metrics on 890 images from the UIEB dataset. Since the EUVP dataset contains a large number of images, we randomly select 500 images from the EUVP$\textbackslash$Paired$\textbackslash$underwater\_dark subset for metric calculation. 
	
	The URPC2022 dataset contains 9000 underwater images, and the target categories include holothurian, echinus, scallop, and starfish. This paper not only uses the dataset to evaluate image enhancement performance, but also relabels the dataset for the training and validation of the YOLOv8 object detection model. During YOLOv8 training, the dataset is divided with a ratio of 0.83:0.17, where 7470 images are used for training and 1530 images are used for validation.
	
	The DUO dataset shows common underwater degradation problems, such as color distortion, low contrast, blur, and complex backgrounds. Its targets are usually small, densely distributed, and easily occluded, which places higher requirements on detection algorithms. This paper uses the DUO dataset as supplementary experimental data. It further verifies the robustness and generalization ability of the proposed method in complex underwater scenes.
	
	\subsection{Experimental Pretreatment}
	
	During the testing stage, models that support arbitrary input sizes process images at their original resolutions. This preserves the spatial resolution of the enhanced results. For models limited by network structure, we first use reflection padding to expand the input image to the required size of the network, and then crop the output back to the original resolution after inference. This operation is only used to satisfy the size compatibility requirements of the network structure and does not introduce additional enhancement information.
	
	\subsection{Experimental Evaluation Criteria}
	
	Image quality metrics include Entropy, UCIQE, UIQM, PSNR, and SSIM. Entropy measures the richness of image information, where a larger value usually indicates that the image contains more information. UCIQE quantitatively evaluates nonuniform color distortion, blur, and low contrast in underwater images. UIQM comprehensively evaluates image color balance, sharpness, and contrast. Peak Signal to Noise Ratio (PSNR) measures the global distortion between the enhanced image and the reference image. SSIM evaluates the similarity between the enhanced image and the reference image from the aspects of brightness, contrast, and structural information.
	
	Real time metrics include Runtime and FPS. Runtime measures the average end to end processing time for a single image. FPS evaluates the real time processing ability of the method, where a larger value indicates a faster running speed.
	
	Object detection metrics include P, R, mAP50, and mAP50:95. P measures how many predicted positive samples are true positive samples. R measures the proportion of real targets that are correctly detected. mAP50 represents the mean average precision of all categories when the IoU threshold is 0.5. mAP50:95 represents the mean average precision when the IoU threshold ranges from 0.5 to 0.95 with a step size of 0.05.
	
	\subsection{Analysis of Image Processing Results}
	
	\subsubsection{Subjective Analysis}
	
    \begin{figure*}[!tbp]

	\begin{minipage}{\textwidth}
		\centering
		\includegraphics[width=.9\textwidth]{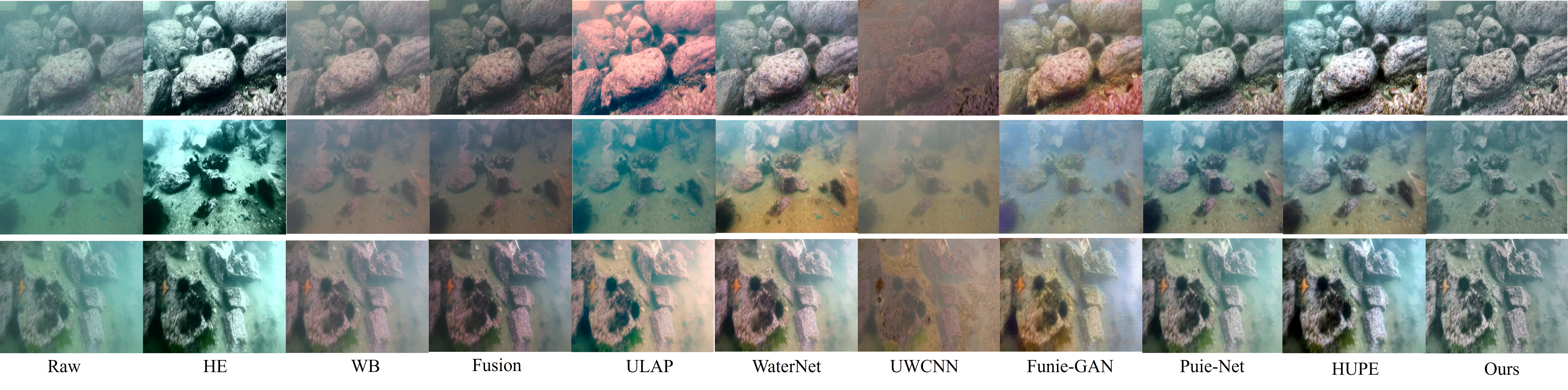}
		\captionsetup{width=\textwidth, justification=justified, singlelinecheck=false, font=small}
		\caption{The visual comparison of cyan cast underwater images. From left to right are: Raw, HE, WB, Fusion \citep{Ancuti2018Fusion}, ULAP \citep{Song2018ULAP}, WaterNet \citep{Li2020UIEB}, UWCNN \citep{Li2020WaterNet}, FUnIE-GAN \citep{Islam2020FUnIEGAN}, PUIE-Net \citep{Fu2022Ushape}, HUPE \citep{Zhang2025HUPE}, and Ours.}
		\label{fig5}
	\end{minipage}

	\vspace{1mm}

	\begin{minipage}{\textwidth}
		\centering
		\includegraphics[width=.9\textwidth]{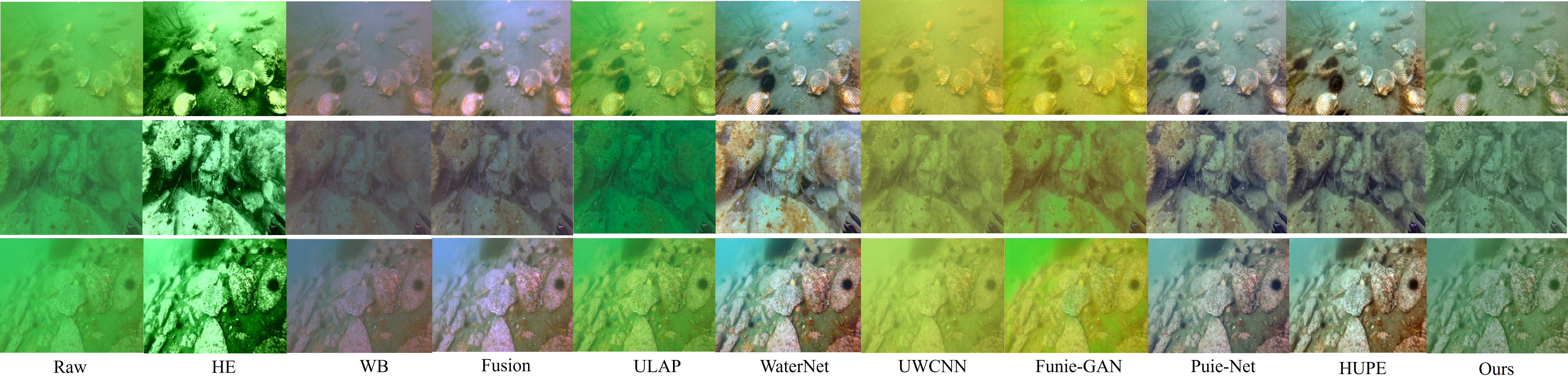}
		\captionsetup{width=\textwidth, justification=justified, singlelinecheck=false, font=small}
		\caption{The visual comparison of green cast underwater images. From left to right are: Raw, HE, WB, Fusion \citep{Ancuti2018Fusion}, ULAP \citep{Song2018ULAP}, WaterNet \citep{Li2020UIEB}, UWCNN \citep{Li2020WaterNet}, FUnIE-GAN \citep{Islam2020FUnIEGAN}, PUIE-Net \citep{Fu2022Ushape}, HUPE \citep{Zhang2025HUPE}, and Ours.}
		\label{fig6}
	\end{minipage}

	\vspace{1mm}

	\begin{minipage}{\textwidth}
		\centering
		\includegraphics[width=.9\textwidth]{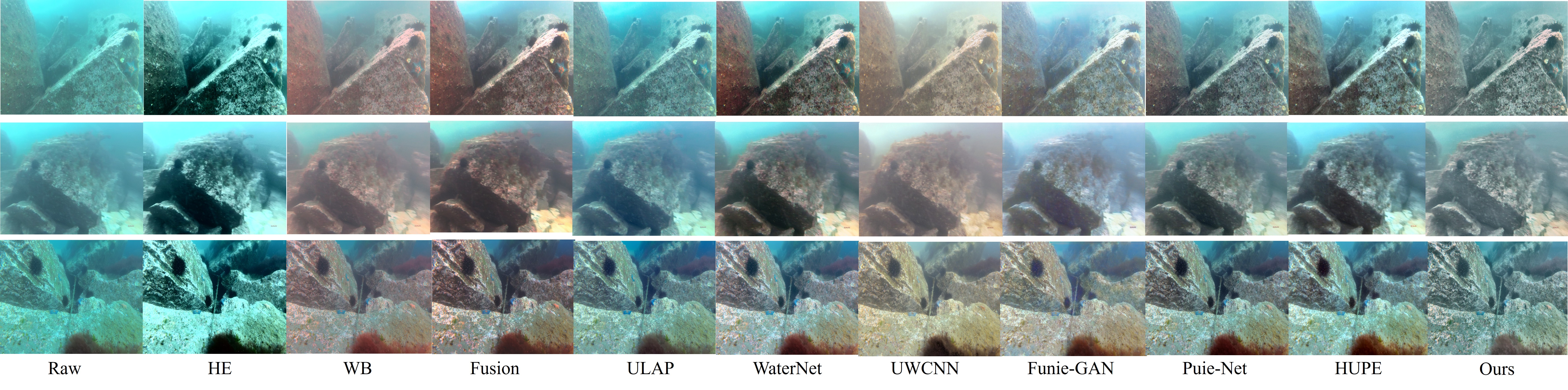}
		\captionsetup{width=\textwidth, justification=justified, singlelinecheck=false, font=small}
		\caption{The visual comparison of blue cast underwater images. From left to right are: Raw, HE, WB, Fusion \citep{Ancuti2018Fusion}, ULAP \citep{Song2018ULAP}, WaterNet \citep{Li2020UIEB}, UWCNN \citep{Li2020WaterNet}, FUnIE-GAN \citep{Islam2020FUnIEGAN}, PUIE-Net \citep{Fu2022Ushape}, HUPE \citep{Zhang2025HUPE}, and Ours.}
		\label{fig7}
	\end{minipage}

	\vspace{1mm}

	\begin{minipage}{\textwidth}
		\centering
		\includegraphics[width=.9\textwidth]{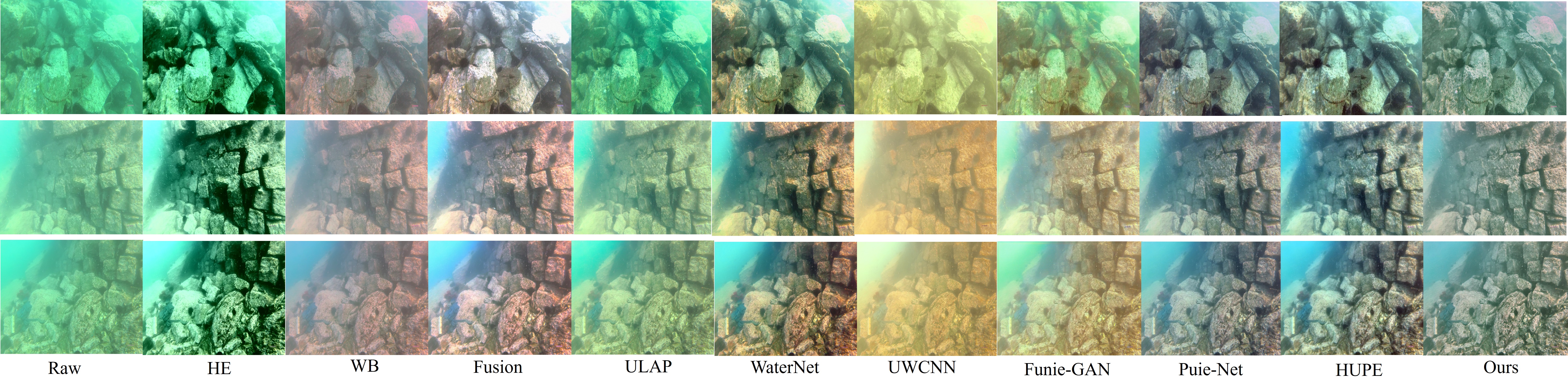}
		\captionsetup{width=\textwidth, justification=justified, singlelinecheck=false, font=small}
		\caption{The visual comparison of exposure-degraded underwater images. From left to right are: Raw, HE, WB, Fusion \citep{Ancuti2018Fusion}, ULAP \citep{Song2018ULAP}, WaterNet \citep{Li2020UIEB}, UWCNN \citep{Li2020WaterNet}, FUnIE-GAN \citep{Islam2020FUnIEGAN}, PUIE-Net \citep{Fu2022Ushape}, HUPE \citep{Zhang2025HUPE}, and Ours.}
		\label{fig8}
	\end{minipage}

\end{figure*}
	
	As shown in Figs.~\ref{fig5}-\ref{fig8}, the enhancement effects of various methods vary under different degradation conditions such as cyan bias, green bias, blue bias, and exposure. Single traditional methods such as HE can improve image contrast to a certain extent, but they may easily cause color distortion or local over enhancement. WB and Fusion improve color correction to some extent, but they still suffer from residual color distortion and insufficient detail restoration. ULAP improves brightness and clarity in some images, but its color stability is limited, and obvious color shifts can still be observed in green cast and exposure degradation scenes. Deep learning-based methods such as WaterNet, UWCNN, FUnIE-GAN, PUIE-Net, and HUPE improve visual quality, but some results still show unnatural colors, uneven brightness, or blurred texture details in complex underwater scenes. Overall, the proposed method shows better stability in different complex underwater environments. It effectively corrects color deviation, improves image brightness and contrast, and produces more natural and clearer visual results.
	
	\subsubsection{Objective Analysis}
	
	This paper selects 30 images from each of four complex underwater scenes in the URPC dataset, including cyan cast, green cast, blue cast, and exposure degradation, to construct the test set. The above objective evaluation metrics are then calculated for the results produced by different methods. Considering the differences among datasets in imaging environment, degradation type, and scene complexity, it is difficult to comprehensively evaluate the effectiveness of the proposed method using only the URPC dataset. Therefore, this paper also conducts experimental analysis on the UIEB and EUVP datasets to evaluate the performance of the proposed method in more diverse scenes. The results are shown in Tables~\ref{tab3} and~\ref{tab4}.
	
	\begin{table*}[width=\textwidth,cols=10,pos=h]
		\caption{No-reference image quality evaluation.}
		\label{tab3}
		\begin{tabular*}{\tblwidth}{@{} C C C C C C C C C C @{}}
			\toprule
			\multirow{2}{*}{Method} 
			& \multicolumn{3}{C}{URPC} 
			& \multicolumn{3}{C}{UIEB} 
			& \multicolumn{3}{C}{EUVP} \\
			\cmidrule(lr){2-4} \cmidrule(lr){5-7} \cmidrule(lr){8-10}
			& Entropy & UCIQE & UIQM 
			& Entropy & UCIQE & UIQM 
			& Entropy & UCIQE & UIQM \\
			\midrule
			Raw       & 6.652 & 0.416 & 0.954 & 6.786 & 0.457 & 1.837 & 7.195 & 0.500 & 2.489 \\
			HE        & \textbf{7.473} & \textbf{0.528} & \textbf{1.596} & 7.298 & \textbf{0.541} & 2.078 & 7.608 & 0.533 & 2.539 \\
			WB        & 6.702 & 0.411 & 1.051 & 6.503 & 0.472 & 1.960 & 7.327 & 0.501 & 2.558 \\
			Fusion    & 7.040 & 0.460 & 1.099 & 6.747 & 0.471 & 1.808 & 7.165 & 0.503 & 2.539 \\
			ULAP      & 6.987 & 0.479 & 1.366 & 6.618 & 0.512 & 1.952 & 7.308 & 0.533 & 2.447 \\
			WaterNet  & 7.438 & 0.505 & 1.525 & 7.418 & 0.517 & 2.211 & 7.592 & 0.522 & 2.556 \\
			UWCNN     & 6.687 & 0.418 & 1.080 & 6.932 & 0.459 & 1.931 & 7.484 & 0.503 & 2.511 \\
			FUnIE-GAN & 7.068 & 0.456 & 1.423 & 7.223 & 0.487 & 2.205 & 7.527 & 0.514 & 2.522 \\
			PUIE-Net  & 7.290 & 0.476 & 1.400 & 7.336 & 0.494 & 2.114 & 7.530 & 0.506 & 2.534 \\
			HUPE      & 7.462 & 0.508 & 1.220 & \textbf{7.489} & 0.531 & 2.142 & \textbf{7.675} & \textbf{0.538} & 2.546 \\
			Ours      & 7.034 & 0.428 & 1.508 & 7.181 & 0.470 & \textbf{2.249} & 7.602 & 0.502 & \textbf{2.576} \\
			\bottomrule
		\end{tabular*}
	\end{table*}
	
	\begin{table}[width=.9\linewidth,cols=5,pos=h]
		\caption{Full-reference image quality evaluation.}
		\label{tab4}
		\begin{tabular*}{\tblwidth}{@{} C C C C C @{}}
			\toprule
			\multirow{2}{*}{Method} 
			& \multicolumn{2}{C}{UIEB} 
			& \multicolumn{2}{C}{EUVP} \\
			\cmidrule(lr){2-3} \cmidrule(lr){4-5}
			& PSNR & SSIM & PSNR & SSIM \\
			\midrule
			Raw       & 28.62 & 0.773 & 28.39 & 0.832 \\
			HE        & 28.11 & 0.771 & 27.96 & 0.699 \\
			WB        & 28.20 & 0.774 & 28.74 & 0.864 \\
			Fusion    & 27.99 & 0.678 & 28.13 & 0.713 \\
			ULAP      & 28.31 & 0.744 & 28.29 & 0.782 \\
			WaterNet  & \textbf{29.35} & \textbf{0.876} & 28.73 & 0.856 \\
			UWCNN     & 28.09 & 0.754 & 28.61 & \textbf{0.880} \\
			FUnIE-GAN & 28.29 & 0.714 & \textbf{28.91} & 0.775 \\
			PUIE-Net  & 28.53 & 0.868 & 28.45 & 0.871 \\
			HUPE      & 29.14 & 0.859 & 28.39 & 0.847 \\
			Ours      & 28.13 & 0.849 & 28.03 & 0.772 \\
			\bottomrule
		\end{tabular*}
	\end{table}

    Due to differences in public implementations, parameter settings, and reproduction conditions, some deep learning methods use an input size of $256 \times 256$ in our experiments. For these methods, we select images with an original resolution of $256 \times 256$ from the EUVP dataset. All methods use the same input size, without additional resizing. Differences in input size may further affect inference efficiency and downstream detection performance. Therefore, we report only the image quality metrics for these methods and present the results as supplementary experiments. TestA contains the same 500 images selected above from the EUVP subset underwater\_dark$\textbackslash$trainA. TestB contains 500 images randomly selected from the EUVP subset underwater\_imagenet$\textbackslash$trainA. The results are shown in Table~\ref{tab5}.
		
	\begin{table*}[width=\textwidth,cols=11,pos=t]
		\caption{Comparison with advanced underwater image enhancement methods.}
		\label{tab5}
		\begin{tabular*}{\tblwidth}{@{} C C C C C C C C C C C @{}}
			\toprule
			\multirow{2}{*}{Method}
			& \multicolumn{5}{C}{TestA}
			& \multicolumn{5}{C}{TestB} \\
			\cmidrule(lr){2-6} \cmidrule(lr){7-11}
			& Entropy & UCIQE & UIQM & PSNR & SSIM
			& Entropy & UCIQE & UIQM & PSNR & SSIM \\
			\midrule
			Ucolor~\citep{Li2021Ucolor}    & 7.562 & 0.515 & 2.547 & 28.61 & \textbf{0.865} & 7.547 & 0.520 & 2.527 & \textbf{28.95} & \textbf{0.873} \\
			PUGAN~\citep{Cong2023PUGAN}     & 7.511 & 0.513 & 2.519 & 28.60 & 0.818 & \textbf{7.655} & \textbf{0.540} & 2.571 & 28.54 & 0.826 \\
			Semi-UIR~\citep{Huang2023SemiUIR} & \textbf{7.711} & \textbf{0.535} & 2.515 & 28.42 & 0.824 & 7.649 & 0.533 & 2.533 & 28.37 & 0.759 \\
			X-CAUNET~\citep{Pramanick2024XCAUNET} & 7.637 & 0.530 & 2.548 & \textbf{28.71} & 0.863 & 7.589 & 0.536 & 2.565 & 28.61 & 0.846 \\
			Ours                  & 7.602 & 0.502 & \textbf{2.576} & 28.03 & 0.772 & 7.528 & 0.491 & \textbf{2.588} & 28.03 & 0.769 \\
			\bottomrule
		\end{tabular*}
	\end{table*}
	
	As shown in Tables~\ref{tab3}-\ref{tab5}, the enhanced images produced by our method outperform the original images on most metrics. This demonstrates that our method effectively improves underwater image quality. The proposed method achieves excellent performance on no reference metrics, especially UIQM, and outperforms other state of the art methods on both the UIEB and EUVP datasets. This indicates that the proposed method can effectively improve the visual perception quality of underwater images. Since this paper focuses more on improving overall visual quality, clarity, and contrast rather than strictly optimizing pixel level reconstruction errors, the proposed method does not achieve the best performance on full reference metrics such as PSNR and SSIM, and some differences still exist between the enhanced images and the reference images.

\begin{figure}
		\centering
		\includegraphics[width=.9\columnwidth]{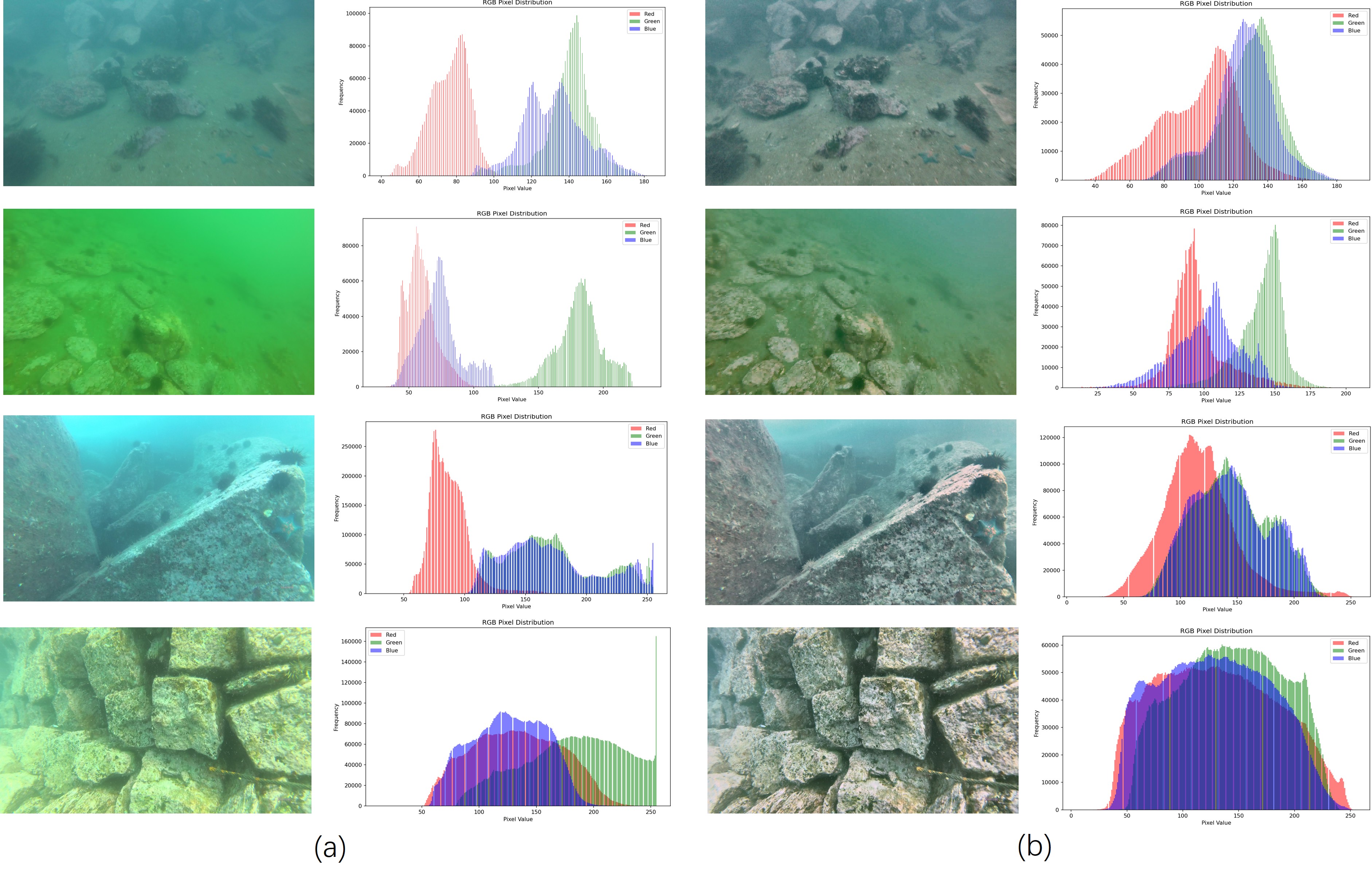}
		\caption{RGB distribution histogram.}\label{fig10}
	\end{figure}

	This paper analyzes the pixel distribution of each color channel before and after enhancement using RGB histograms, which provides a more intuitive evaluation of color correction and dynamic range in different complex underwater scenes. The results are shown in Fig.~\ref{fig10}.
	
	 It can be observed that the peaks of different color channels in group (b) become more concentrated, and the separation between channel intervals is clearly improved. Compared with the scattered histogram distribution or excessive overflow of a single channel in the original images of group (a), the enhanced results show obvious improvement. These results indicate that the proposed method can effectively optimize image color distribution, improve the utilization of color information, reduce pixel redundancy, and enhance overall color balance.
	
	\subsection{Real-time Analysis}
	
	To further evaluate the runtime efficiency of different methods, we select images with five representative resolutions from URPC2022, UIEB, and EUVP: $256 \times 256$, $640 \times 480$, $1280 \times 720$, $1920 \times 1080$, and $3840 \times 2160$. All images are converted to JPG format to ensure consistent testing conditions. The average runtime per image and the corresponding FPS are reported in Table~\ref{tab6}. HUPE requires additional prior information during inference, and its complete processing pipeline cannot be fully reproduced under the unified timing environment adopted in this study. Therefore, HUPE is excluded from the runtime comparison.
	
	\begin{table*}[width=\textwidth,cols=8,pos=t]
		\caption{Runtime and model parameter comparison.}
		\label{tab6}
		\begin{tabular*}{\tblwidth}{@{} C C C C C C C C C C C C @{}}
			\toprule
			\multirow{2}{*}{Method}
			& \multicolumn{2}{C}{$256 \times 256$}
			& \multicolumn{2}{C}{$640 \times 480$}
			& \multicolumn{2}{C}{$1280 \times 720$}
            & \multicolumn{2}{C}{$1920 \times 1080$}
            & \multicolumn{2}{C}{$3840 \times 2160$}
			& \multirow{2}{*}{Parameters} \\
			\cmidrule(lr){2-3} \cmidrule(lr){4-5} \cmidrule(lr){6-7} \cmidrule(lr){8-9} \cmidrule(lr){10-11}
			& Time (s) & FPS & Time (s) & FPS & Time (s) & FPS & Time (s) & FPS & Time (s) & FPS \\
			\midrule
			WaterNet  & 0.085  & 11.779 & 0.243 & 4.115 & 0.622 & 1.608 &1.443 & 0.693 & 46.278 & 0.024 & 1.09M \\
			UWCNN     & 0.216  & 4.630 & 0.395 & 2.534 & 0.883 & 1.132  &  1.892& 0.529& 7.465& 1.134&0.04M \\
			FUnIE-GAN & 0.018  & 54.688 & 0.044 & 22.386 & 0.083 & 12.004 & 0.137 & 7.292 & 0.597 & 1.676 & 7.02M \\
			PUIE-Net  & 0.042  & 23.726 & 0.076 & 13.191 & 0.187 & 5.335 & 0.394 & 2.539 & 14.364 & 0.070 &  16.12M \\
			Ours      & \textbf{0.009}  & \textbf{106.693} & \textbf{0.020} & \textbf{50.900} & \textbf{0.048} & \textbf{20.762} & \textbf{0.108} & \textbf{9.218} & \textbf{0.388} & \textbf{2.579} & -- \\
			\bottomrule
		\end{tabular*}
	\end{table*}
	
	The results show that our method adapts well to different input resolutions. It maintains high processing efficiency, especially for high-resolution images.
	
	\subsection{Feature Analysis}
	
	The above objective image metrics quantitatively describe enhancement results from the perspectives of global statistical characteristics and reference consistency. Although these metrics can reflect the overall quality change of images, they still lack intuitive evaluation of whether target regions are effectively highlighted, background interference is suppressed, and local textures, edges, and structural information are sufficiently preserved. Especially in underwater image enhancement tasks, enhancement results should not only improve visual quality, but also provide more reliable inputs for subsequent object detection, feature extraction, and image matching.
	
	Therefore, this paper uses saliency detection and feature matching experiments to qualitatively analyze the enhancement results from two perspectives: target saliency representation and local feature preservation.
	
	\begin{figure}
		\centering
		\includegraphics[width=.9\columnwidth]{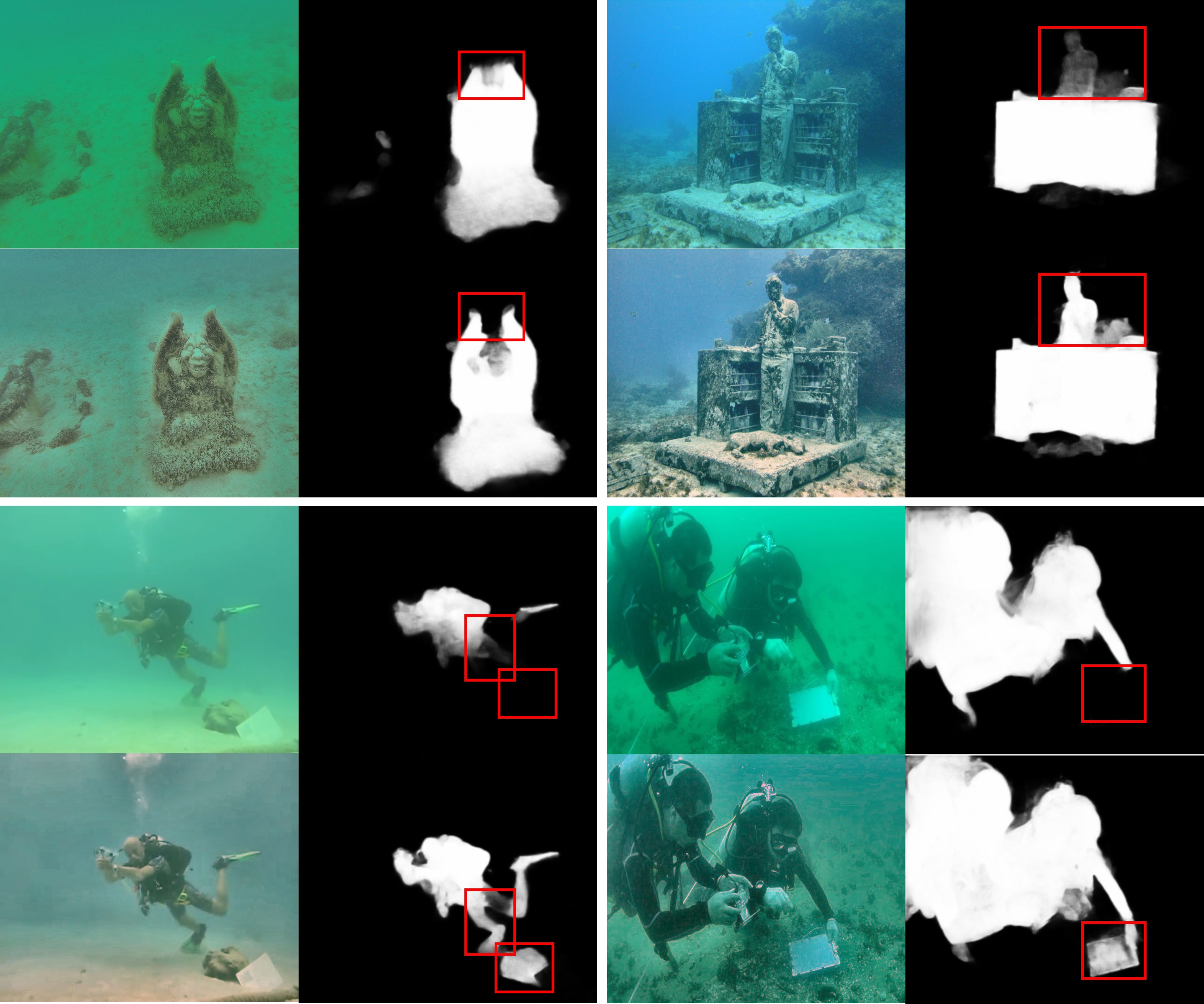}
		\caption{Significance detection results of U2NET~\citep{Qin2020U2Net}. The upper row shows the original input results, while the lower row displays the results processed by our method.}\label{fig11}
	\end{figure}
	
	\begin{figure}
		\centering
		\includegraphics[width=.9\columnwidth]{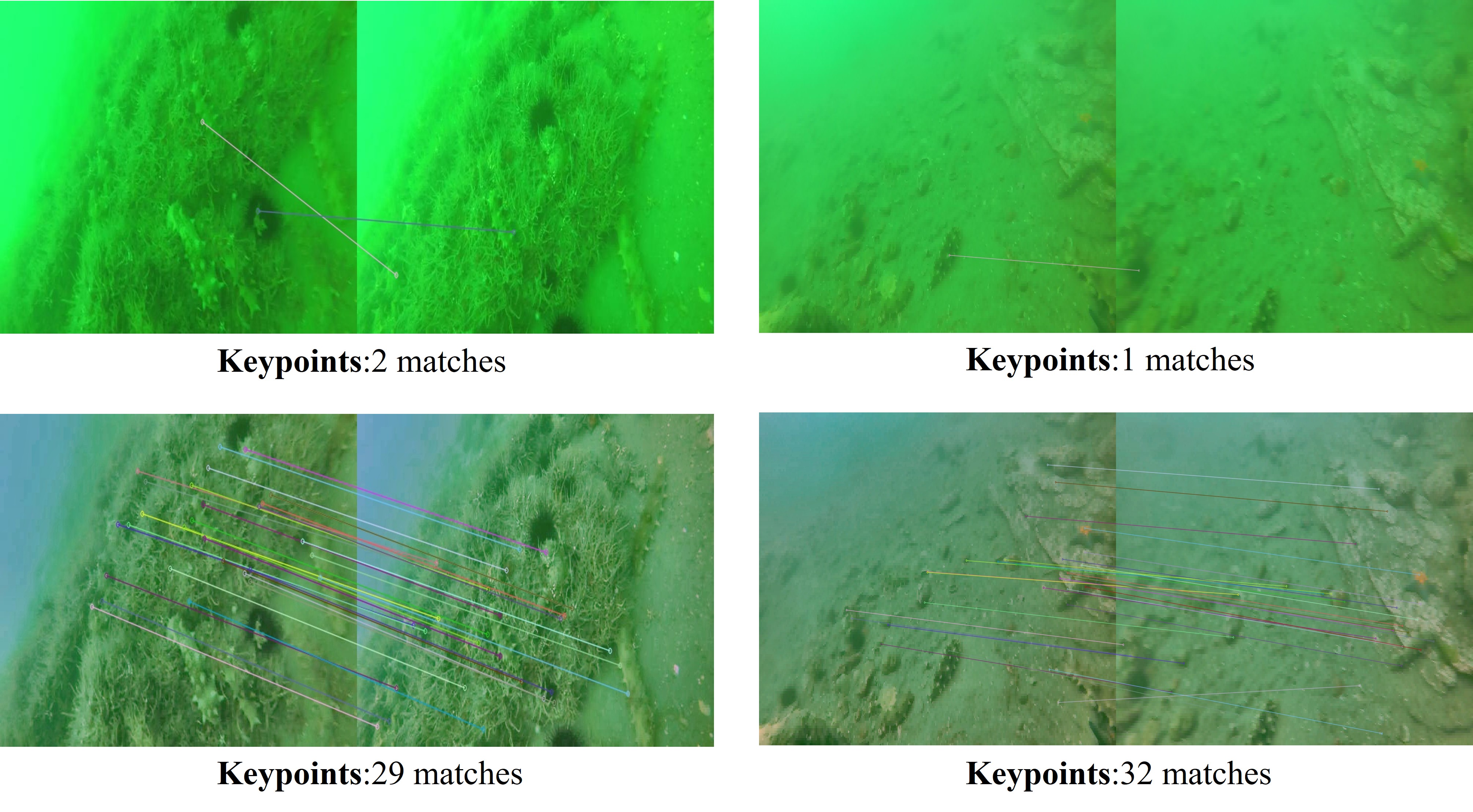}
		\caption{Feature matching results ~\citep{Lowe2004SIFT}. The upper row shows the original input results, while the lower row displays the results processed by our method.}\label{fig12}
	\end{figure}
	
	The above experiments, including image quality evaluation, real time analysis, saliency detection, and feature matching, verify the effectiveness of the proposed method from three aspects: image quality, computational efficiency, and feature representation ability. These results provide a relatively comprehensive evaluation of enhancement performance, but their evaluation focus is still mainly at the image level. To further show the application value of enhanced results in practical vision tasks, this paper introduces object detection in the following experiments. The changes in detection performance are used to evaluate how different enhancement methods support downstream tasks.
	
	\subsection{Analysis of Target Detection Results}
	
	\subsubsection{YOLOv8 Detection Results on URPC}
	
	\begin{table}[width=.9\linewidth,cols=5,pos=h]
		\caption{Object detection results on the URPC dataset.}
		\label{tab7}
		\begin{tabular*}{\tblwidth}{@{} C C C C C @{}}
			\toprule
			Method & P & R & mAP50 & mAP50:95 \\
			\midrule
			Raw       & \textbf{0.830} & 0.705 & 0.794 & 0.445 \\
			HE        & 0.807 & 0.680 & 0.769 & 0.419 \\
			WB        & 0.813 & 0.716 & 0.803 & 0.449 \\
			Fusion    & 0.797 & 0.696 & 0.776 & 0.429 \\
			ULAP      & 0.815 & 0.711 & 0.792 & 0.442 \\
			WaterNet  & 0.820 & 0.717 & 0.805 & 0.447 \\
			UWCNN     & 0.804 & 0.701 & 0.782 & 0.437 \\
			FUnIE-GAN & 0.806 & 0.692 & 0.776 & 0.426 \\
			PUIE-Net  & 0.815 & 0.723 & 0.807 & 0.451 \\
			HUPE      & 0.819 & 0.732 & 0.809 & 0.454 \\
			Ours      & 0.818 & \textbf{0.735} & \textbf{0.815} & \textbf{0.456} \\
			\bottomrule
		\end{tabular*}
	\end{table}

	Comprehensive analysis: According to Tables~\ref{tab3}-\ref{tab4} and Table~\ref{tab7}, there is no simple positive correlation between image quality evaluation metrics and object detection performance. Entropy increases when the image contains richer gray level distribution and more detailed information, but it may also reflect more noise and background interference. UCIQE and UIQM improve when the image obtains better color, contrast, and clarity, but higher values do not always mean that the target features required by the detector are also enhanced. PSNR and SSIM increase when the enhanced image is closer to the reference image in pixel values and structural distribution, which indicates smaller reconstruction errors and higher similarity. However, these full reference metrics mainly describe reconstruction consistency and have limited ability to reflect target saliency and detection usability. Therefore, some methods achieve good objective evaluation metrics, but do not obtain ideal results in the object detection task. In comparison, the proposed method not only improves image quality, but also promotes underwater object detection performance. It improves mAP50 by 2.1\%, and achieves the best results in recall and mAP50-95, reaching 0.735 and 0.456, respectively.
	
	\subsubsection{Visualization Results}
	
	This paper visualizes the test results in four underwater environments: cyan-dominated, green-dominated, blue-dominated, and exposed conditions. The following images demonstrate the detection performance and corresponding confidence levels for sea cucumbers, sea urchins, scallops, and starfish.
	
	\begin{figure}
		\centering
		\includegraphics[width=.9\columnwidth]{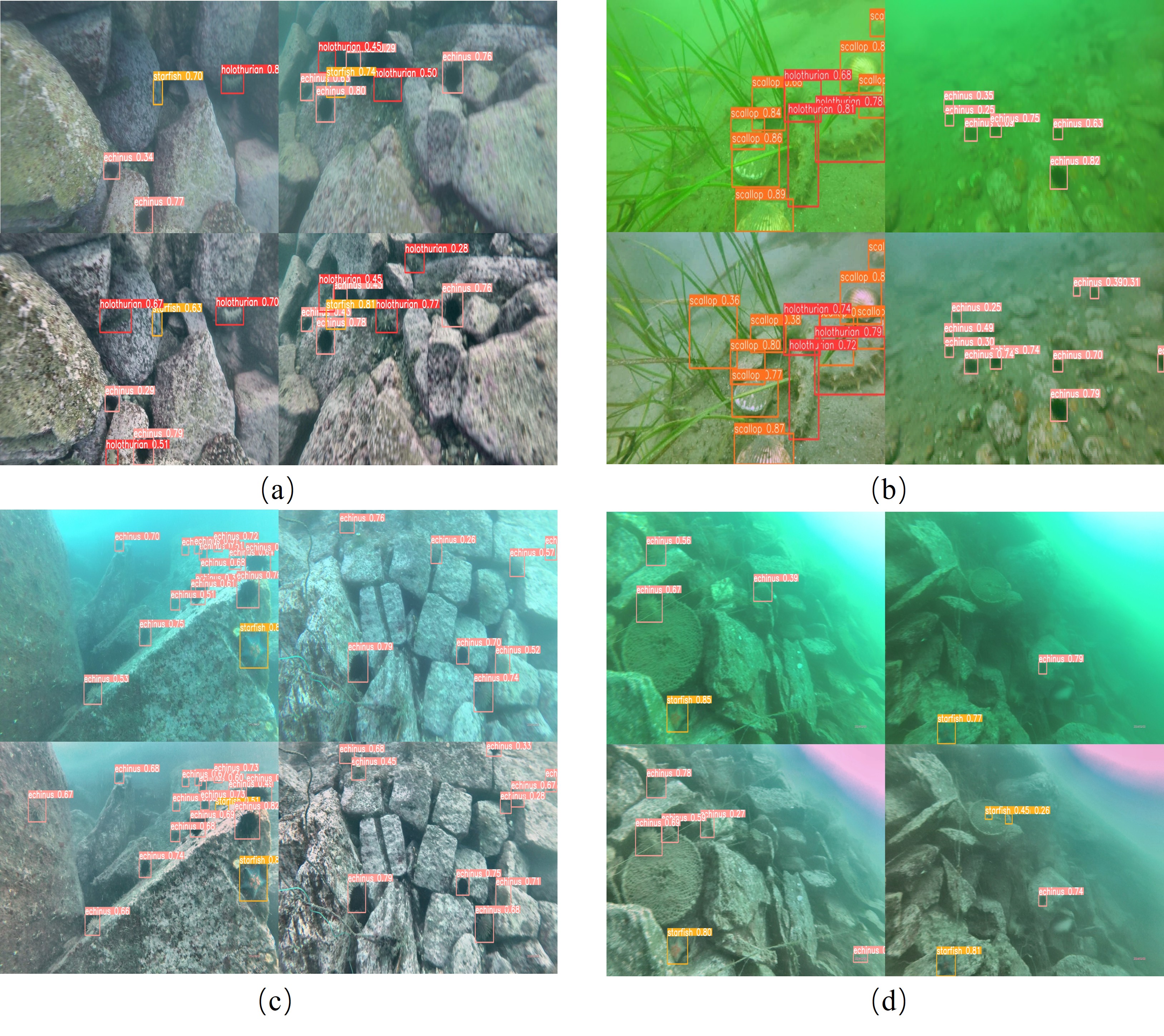}
		\caption{Detection results under different underwater environments. (a) Cyan cast underwater environment; (b) Green cast underwater environment; (c) Blue cast underwater environment; and (d) Exposure degraded underwater environment. The upper row of each group shows the detection results on the raw images, while the lower row presents the detection results after image enhancement.}\label{fig13}
	\end{figure}
	\begin{figure}
		\centering
		\includegraphics[width=.9\columnwidth]{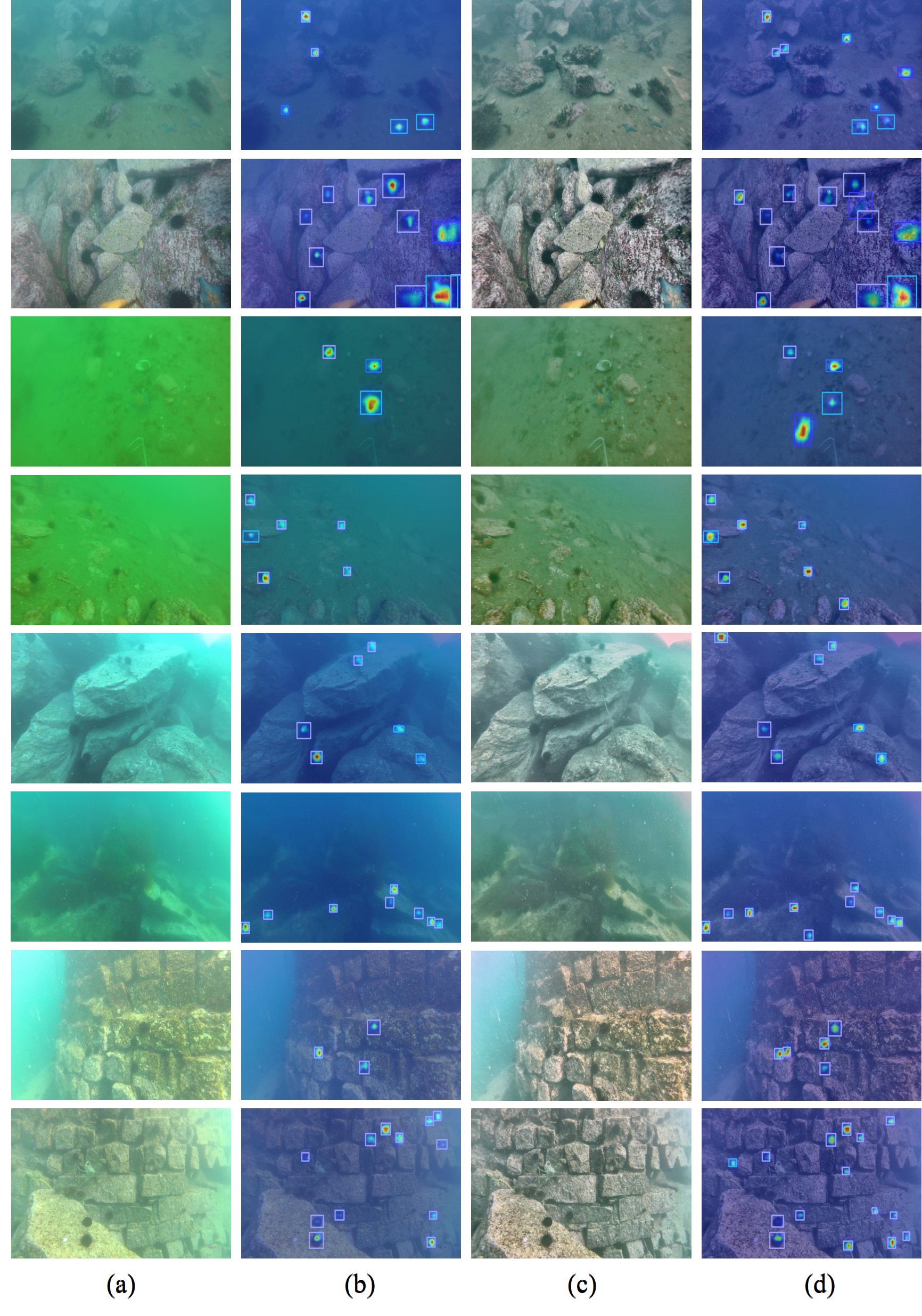}
		\caption{Thermal maps under different environmental conditions. (a) Original image; (b) Thermal map of the original image; (c) Post-processing image; (d) Thermal map of the post-processing image.}\label{fig14}
	\end{figure}

	As shown in Fig.~\ref{fig13}, the original image in the cyan cast scene shows a cold color tone, and the layers between the targets and the rock background are not clear enough. After processing by the proposed method, the image color becomes more natural, and the target contours and rock textures are enhanced, making targets such as echinus and holothurian easier to distinguish. In the green cast scene, severe color distortion reduces the difference between the targets and the background. After processing, the color distortion is improved, the local contrast is enhanced, and the target regions become more prominent. In the blue cast scene, the target edges and local textures become clearer after processing, which improves the recognition of densely distributed echinus. In the abnormal exposure scene, the original image suffers from uneven brightness distribution and local detail loss. After processing, some image details are restored, which improves the object detection performance under complex lighting conditions.
	
	The proposed method shows good enhancement effects in four typical complex underwater environments. It effectively reduces visual degradation caused by color distortion and abnormal exposure, improves the distinction between targets and backgrounds, and enables the detection model to obtain clearer target feature representation. The results show that the proposed method detects more targets, achieves more accurate localization, and maintains more stable confidence scores after enhancement.

	As shown in Fig.~\ref{fig14}, the proposed method shows stronger feature response and localization ability in complex underwater backgrounds. Compared with the detection results of the original images, the proposed method produces more concentrated heatmap responses on target regions, which reduces missed detections and false detections of small underwater targets.
	
	\begin{figure}
		\centering
		\includegraphics[width=.9\columnwidth]{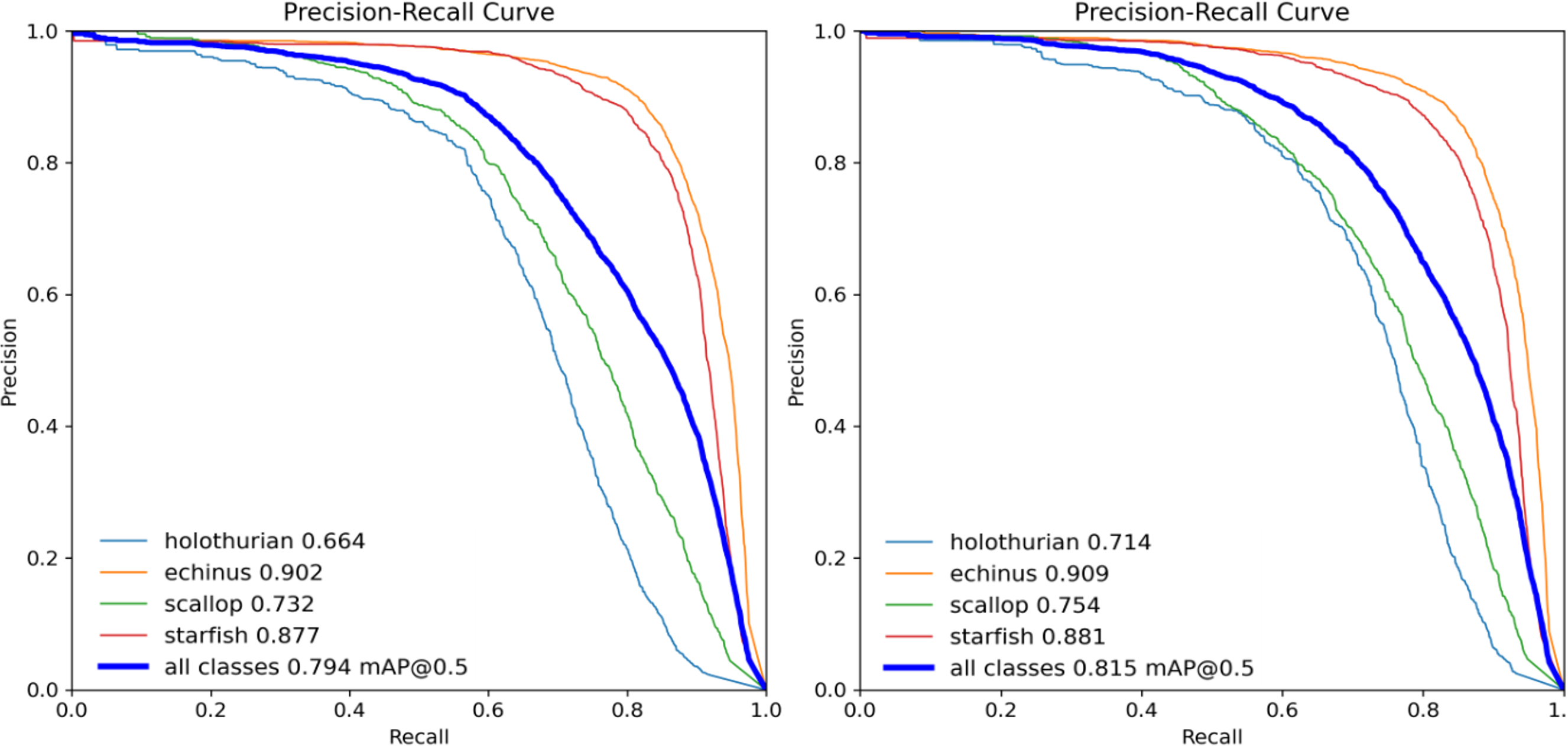}
		\caption{Comparison of P-R curves.}\label{fig15}
	\end{figure}
	
	Fig.~\ref{fig15} shows that the P-R curve of the enhanced images is closer to the upper right corner and keeps higher precision in most recall ranges. This indicates that the proposed method improves target feature representation and reduces background interference, which helps the detector obtain better detection results.
	
	\subsubsection{Analysis of the Reasons for Improved Detection Performance}
	
	\begin{figure}
		\centering
		\includegraphics[width=.9\columnwidth]{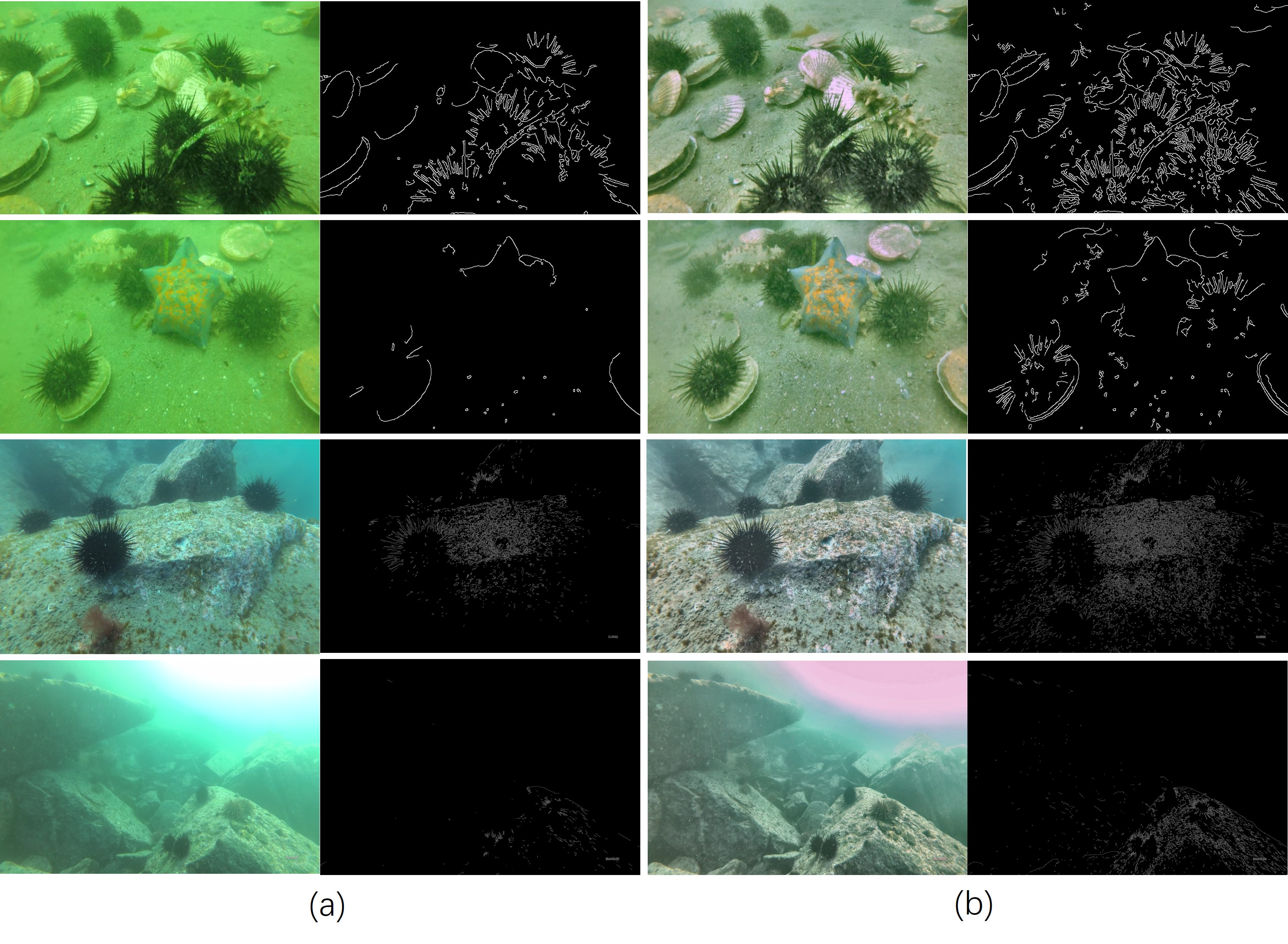}
		\caption{Canny edge detection.}\label{fig16}
	\end{figure}
	
	This paper uses the Canny operator to compare edge detection results between the original images and the enhanced images. Fig.~\ref{fig16} shows that the proposed method makes target edges clearer. Structures such as echinus spines, shell contours, and small target boundaries become more obvious, indicating that the proposed method enhances target edge information and helps improve Recall and mAP. However, the proposed method also enhances some background textures, which may introduce certain false detections and prevent Precision from achieving the best result.

    \subsubsection{Ablation Experiment}

        \begin{table*}[width=\textwidth,cols=9,pos=t]
        \caption{Ablation experiments.}
        \label{tab:ablation}
        \small
        \setlength{\tabcolsep}{3pt}
        \renewcommand{\arraystretch}{1.08}
        \begin{tabular*}{\tblwidth}{@{\extracolsep{\fill}} c c c c c c c c c @{}}
            \toprule
            \multirow{2}{*}{Type} 
            & \multirow{2}{*}{Experimental setting} 
            & \multicolumn{3}{c}{Image quality}
            & \multicolumn{4}{c}{Object detection results} \\
            \cmidrule(lr){3-5} \cmidrule(lr){6-9}
            & & Entropy & UCIQE & UIQM & P & R & mAP50 & mAP50:95 \\
            \midrule
            
            \multirow{5}{*}{Internal ablation}
            & w/o A1 & \textbf{7.210} & \textbf{0.451} & 1.530 & 0.829 & 0.722 & 0.809 & 0.452 \\
            & w/o A2 & 7.037 & 0.428 & 1.424 & 0.823 & 0.731 & 0.811 & 0.453 \\
            & w/o A3 & 6.965 & 0.420 & 1.515 & 0.818 & 0.738 & 0.813 & 0.454 \\
            & w/o B1 & 7.027 & 0.431 & 1.484 & 0.828 & 0.715 & 0.804 & 0.449 \\
            & w/o B2 & 6.506 & 0.385 & 1.039 & 0.822 & 0.729 & 0.812 & \textbf{0.456} \\
            \midrule
            
            \multirow{13}{*}{Structural ablation}
            & Only A & 5.993 & 0.355 & 1.278 & 0.813 & 0.697 & 0.790 & 0.439 \\
            & Only B & 7.206 & 0.452 & 1.563 & 0.826 & 0.720 & 0.804 & 0.453 \\
            & A$\rightarrow$B & 6.976 & 0.431 & 1.912 & 0.818 & 0.703 & 0.793 & 0.441 \\
            & B$\rightarrow$A & 6.179 & 0.357 & 1.352 & 0.810 & 0.697 & 0.783 & 0.433 \\
            & A//B ($\partial=0.10$) & 7.126 & 0.440 & \textbf{1.536} & 0.822 & 0.728 & 0.809 & 0.452 \\
            & A//B ($\partial=0.15$) & 7.081 & 0.434 & 1.523 & 0.820 & 0.727 & 0.811 & 0.453 \\
            & A//B ($\partial=0.20$) & 7.034 & 0.428 & 1.508 & 0.818 & \textbf{0.735} & \textbf{0.815} & \textbf{0.456} \\
            & A//B ($\partial=0.25$) & 6.986 & 0.422 & 1.494 & 0.824 & 0.728 & 0.811 & 0.453 \\
            & A//B ($\partial=0.30$) & 6.936 & 0.417 & 1.480 & 0.825 & 0.723 & 0.809 & 0.452 \\
            & A//B ($\partial=0.35$) & 6.884 & 0.411 & 1.466 & \textbf{0.833} & 0.723 & 0.809 & 0.454 \\
            & A//B ($\partial=0.50$) & 6.718 & 0.396 & 1.424 & 0.825 & 0.720 & 0.810 & 0.453 \\
            & A//B ($\partial=0.65$) & 6.533 & 0.382 & 1.381 & 0.815 & 0.720 & 0.803 & 0.450 \\
            & A//B ($\partial=0.80$) & 6.319 & 0.370 & 1.336 & 0.816 & 0.713 & 0.799 & 0.444 \\
            \bottomrule
        \end{tabular*}
    \end{table*}

    The previous analysis shows that image quality metrics do not strictly correspond to object detection performance. Higher entropy may result from amplified underwater noise and enhanced information in non target regions. Higher UIQM and UCIQE values reflect improvements in color, contrast, and clarity. However, these metrics cannot show whether the improvements mainly occur in target regions or backgrounds. If fusion weights rely only on image quality metrics, the parameters may favor global visual enhancement. This choice may not preserve and highlight useful features in target regions. Therefore, we adopt a task based evaluation strategy. We select the final parameters according to detection performance, while keeping image quality stable.

    The internal ablation study in Table~\ref{tab:ablation} shows that removing any module reduces detection performance. This result confirms that each module contributes to the final detection performance. Removing the B1 module reduces mAP50 from 0.815 to 0.804, which gives the largest drop. This result shows that B1 plays an important role in target feature representation. Removing the other modules causes smaller drops in detection performance, but none reaches the full model. These results indicate complementary and collaborative effects among the modules.

    The structural ablation study in Table~\ref{tab:ablation} shows that neither single branch structures nor cascaded structures achieve the best performance. Branch A focuses on detail enhancement, while Branch B focuses on color restoration. Used alone, each branch cannot fully preserve target details and color consistency. Although A followed by B and B followed by A combine both types of information, the later operation may weaken some details or color information. In contrast, the parallel structure preserves useful information from both branches. The fusion process combines their complementary strengths. Therefore, the parallel structure achieves better detection performance than single branch and cascade structures.

    \begin{figure}
		\centering
		\includegraphics[width=.9\columnwidth]{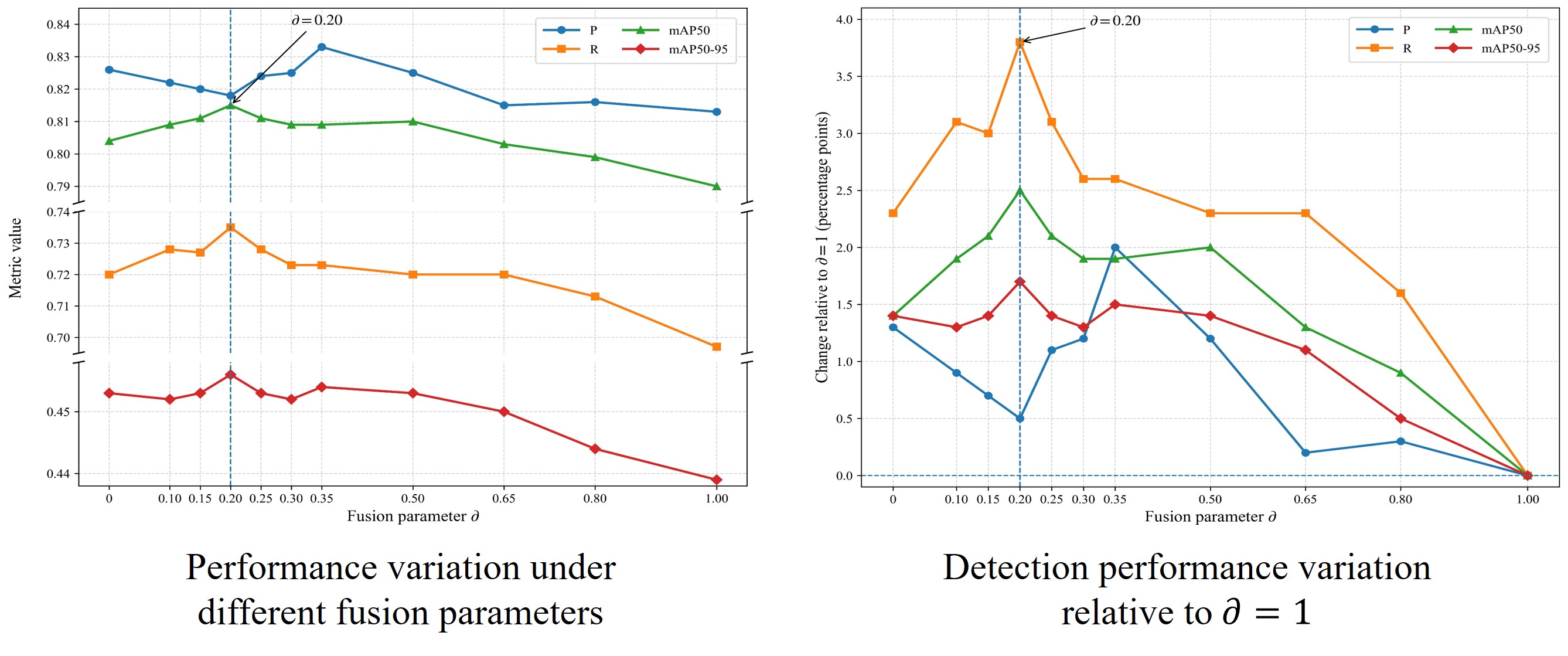}
		\caption{Sensitivity analysis of the fusion parameter.}\label{fig17}
	\end{figure}

    As shown in Table~\ref{tab:ablation} and Fig.~\ref{fig17}, ($\partial$ = 0.20) lies within a stable high performance region. At this setting, Recall, mAP50, and mAP50-95 reach strong levels, while image quality metrics remain stable. This result shows that the setting balances visual enhancement quality and downstream detection performance. Therefore, we use a fusion weight of ($\partial$ = 0.20) in all experiments.

	\subsubsection{Comparative Analysis Under Different Loss Functions}
	
	After confirming the overall effectiveness of the proposed enhancement method, we further analyze the detection stability of enhanced images. We compare several representative bounding box regression losses, including GIoU \cite{Rezatofighi2019GIoU}, DIoU and CIoU \cite{Zheng2020DIoU}, EIoU \cite{Zhang2022EIoU}, SIoU \cite{Gevorgyan2022SIoU}, and WIoU \cite{Tong2023WiseIoU}. We keep the detection network, training settings, and dataset split unchanged, and only replace the loss function. Table~\ref{tab8} presents the results.
	
	\begin{table}[width=.9\linewidth,cols=5,pos=h]
		\caption{Comparison of different IoU loss functions.}
		\label{tab8}
		\begin{tabular*}{\tblwidth}{@{} C C C C C @{}}
			\toprule
			Loss & P & R & mAP50 & mAP50:95 \\
			\midrule
			GIoU & 0.814 & 0.730 & 0.810 & 0.453 \\
			DIoU & 0.821 & 0.734 & 0.811 & 0.452 \\
			CIoU & 0.818 & 0.735 & 0.815 & \textbf{0.456} \\
			SIoU & 0.824 & 0.727 & 0.813 & 0.455 \\
			EIoU & 0.820 & 0.724 & 0.806 & 0.450 \\
			WIoU v1 & 0.816 & \textbf{0.737} & 0.810 & 0.455 \\
			WIoU v2 ($\gamma=0.5$) & 0.821 & 0.735 & \textbf{0.816} & 0.454 \\
			WIoU v3 ($\alpha=1.9,\ \delta=3$) & \textbf{0.826} & 0.733 & 0.814 & \textbf{0.456} \\
			WIoU v3 ($\alpha=1.6,\ \delta=4$) & 0.815 & 0.731 & 0.812 & 0.453 \\
			WIoU v3 ($\alpha=1.4,\ \delta=5$) & \textbf{0.826} & 0.729 & 0.813 & 0.454 \\
			\bottomrule
		\end{tabular*}
	\end{table}
	
	Based on the edge and saliency results, our method strengthens object contours and texture details. It also improves the separation between targets and backgrounds, reducing geometric errors between predicted and ground truth boxes. Therefore, CIoU can better use its center distance and aspect ratio constraints. After enhancement, some blurry targets become more informative samples. WIoU reduces the influence of abnormal samples through dynamic sample weighting. It also focuses more on medium quality samples with greater optimization value. As a result, WIoU achieves better overall detection performance.
	
	\subsubsection{YOLOv8 Detection Results on DUO}
	
	\begin{table}[width=.9\linewidth,cols=5,pos=h]
		\caption{Object detection results on the DUO dataset.}
		\label{tab9}
		\begin{tabular*}{\tblwidth}{@{} C C C C C @{}}
			\toprule
			Method & P & R & mAP50 & mAP50:95 \\
			\midrule
			Raw       & 0.822 & 0.727 & 0.803 & 0.602 \\
			HE        & 0.835 & 0.668 & 0.770 & 0.551 \\
			WB        & 0.836 & 0.718 & 0.806 & 0.600 \\
			Fusion    & 0.787 & 0.680 & 0.763 & 0.532 \\
			ULAP      & 0.798 & 0.729 & 0.798 & 0.590 \\
			WaterNet  & 0.830 & 0.712 & 0.801 & 0.587 \\
			UWCNN     & 0.841 & 0.699 & 0.794 & 0.585 \\
			FUnIE-GAN & 0.820 & 0.667 & 0.773 & 0.560 \\
			PUIE-Net  & \textbf{0.853} & 0.715 & \textbf{0.812} & \textbf{0.604} \\
			HUPE      & 0.829 & 0.725 & 0.805 & 0.598 \\
			Ours      & 0.817 & \textbf{0.735} & 0.811 & 0.603 \\
			\bottomrule
		\end{tabular*}
	\end{table}
	
	The results are shown in Table~\ref{tab9}, the proposed method achieves the highest recall of 0.735, while mAP50 and mAP50-95 reach 0.811 and 0.603, respectively, which are the second best results. These results indicate that the proposed method has good generalization ability in different underwater scenes.

    \subsection{Failure Case Analysis}

    \begin{figure}
		\centering
		\includegraphics[width=.9\columnwidth]{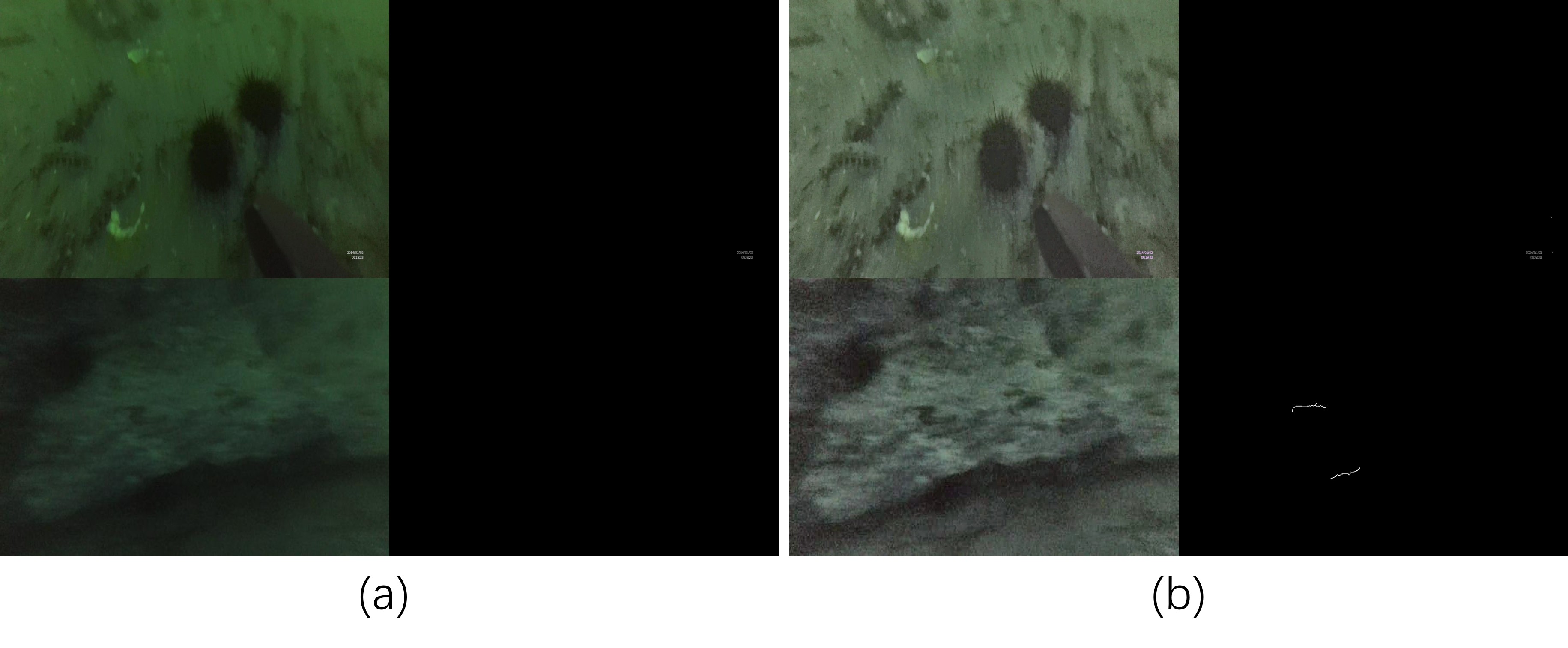}
		\caption{Typical failure cases in extremely dark and blurry scenes.}\label{fig18}
	\end{figure}

    Although our method improves image brightness, color quality, and object edges in most underwater scenes, it performs less effectively in extremely dark and blurry conditions. Fig.~\ref{fig18} presents typical failure cases. The enhancement process improves overall brightness and local contrast, making the scenes easier to recognize. However, the Canny edge maps show that the method does not recover useful target information.

    Two main factors cause these results. Under extremely low light and blur, the original contrast between the target and the background is weak. Some texture and contour information has already been severely lost. The detail enhancement branch highlights high frequency responses through log-domain illumination correction and local contrast enhancement. However, it can only strengthen the structural information that remains in the input image. It cannot reconstruct target edges that have been severely lost. The color restoration branch uses a statistical gain adjustment strategy guided by brightness. In extremely dark regions, low brightness weights suppress excessive color compensation. This reduces the risk of background noise amplification and local color overcorrection. However, it also limits the recovery of targets in dark regions. When the target area is dim and blurry, the color restoration branch cannot further improve local visibility.

    For extremely low light and severe blur, future work will introduce mechanisms that sense the degree of degradation and fusion strategies that adapt to different regions. These improvements will enhance the adaptability of the method to complex underwater environments.
	
	\section{Conclusion}\label{Conclusion}

    This paper addresses the issues of detail degradation and color shifts in underwater images, which lead to reduced performance in object detection, by proposing an efficient dual-branch underwater image enhancement framework. By constructing a detail enhancement branch and a color restoration branch, the framework achieves joint optimization of detail and color information in underwater images, thereby improving image quality while enhancing the suitability of the enhanced results for object detection tasks.
    
	Experimental results show that the proposed method achieves good overall performance on multiple underwater image datasets. It performs better than the original images on most image quality metrics. It also achieves excellent UIQM results on both the UIEB and EUVP datasets and performs better than other state of the art methods. At a resolution of $256 \times 256$, the average processing time per image is only 0.009s. In the object detection task, the enhanced images achieve the best results on Recall, mAP50, and mAP50-95, with mAP50 improved by 2.1\% compared with the baseline. The results on the DUO dataset further show that the proposed method improves underwater object detection performance in different scenes. These results indicate that the proposed method achieves a good balance among image enhancement quality, processing efficiency, and object detection performance.

	\printcredits
    
	\section*{Declaration of competing interest}
    The authors declare that they have no known competing financial interests or personal relationships that could have appeared to influence the work reported in this paper.

    \section*{Data availability}
    Data will be made available on request.
	\bibliographystyle{cas-model2-names}
	
	\bibliography{cas-refs}
	
	
	
\end{document}